%% file: access.tex
\documentclass{ieeeaccess}
\usepackage{cite}
\usepackage{amsmath,amssymb,amsfonts}
\usepackage{graphicx}
\usepackage{hyperref}
\usepackage{textcomp}
\usepackage{booktabs} 
\usepackage{stfloats}
\usepackage{algorithm}
\usepackage{algpseudocode}
\usepackage{subcaption} 
\usepackage{soul} 
\usepackage{multirow}
\usepackage{longtable}
\usepackage{adjustbox}
\usepackage{pdflscape}
\usepackage{bm}
\usepackage{ragged2e}
\makeatletter
\AtBeginDocument{\DeclareMathVersion{bold}
\SetSymbolFont{operators}{bold}{T1}{times}{b}{n}
\SetSymbolFont{NewLetters}{bold}{T1}{times}{b}{it}
\SetMathAlphabet{\mathrm}{bold}{T1}{times}{b}{n}
\SetMathAlphabet{\mathit}{bold}{T1}{times}{b}{it}
\SetMathAlphabet{\mathbf}{bold}{T1}{times}{b}{n}
\SetMathAlphabet{\mathtt}{bold}{OT1}{pcr}{b}{n}
\SetSymbolFont{symbols}{bold}{OMS}{cmsy}{b}{n}
\renewcommand\boldmath{\@nomath\boldmath\mathversion{bold}}}
\makeatother
\usepackage{rotating}
\usepackage{tabularx} 

\def\BibTeX{{\rm B\kern-.05em{\sc i\kern-.025em b}\kern-.08em
    T\kern-.1667em\lower.7ex\hbox{E}\kern-.125emX}}

\begin{document}
\history{This article has been accepted for publication in IEEE Access. This is the author's version which has not been fully edited and
content may change prior to final publication. Citation information: DOI 10.1109/ACCESS.2025.3635541}
\doi{10.1109/ACCESS.2025.3635541}

\title{The Impact of Feature Scaling In Machine Learning:
Effects on Regression and Classification Tasks}
\author{\uppercase{João Manoel Herrera Pinheiro}\authorrefmark{1}, \uppercase{Suzana Vilas Boas de Oliveira}\authorrefmark{2}, \uppercase{Thiago Henrique Segreto Silva}\authorrefmark{1}, \uppercase{Pedro Antonio Rabelo Saraiva}\authorrefmark{1}, \uppercase{Enzo Ferreira de Souza}\authorrefmark{1}, \uppercase{Ricardo V. Godoy}\authorrefmark{1}, \uppercase{Leonardo André Ambrosio}\authorrefmark{2}, \uppercase{Marcelo Becker}\authorrefmark{1}}

\address[1]{Department of Mechanical Engineering, University of São Paulo,13566-590, São Paulo, Brazil}
\address[2]{Department of Electrical and Computer Engineering, University of São Paulo,13566-590, São Paulo, Brazil}
\tfootnote{This work was supported by the Petróleo Brasileiro S/A - Petrobras, using resources from the R\&D clause of the ANP, in partnership with the Universidade de São Paulo (USP) and the Fundação de Apoio à Física e à Química (FAFQ), under Cooperation Agreement No. 2023/00016-6 and 2023/00013-7, Coordenação de Aperfeiçoamento de Pessoal de Nível Superior (CAPES), grant nº 88887.992906/2024-00, and National Council for Scientific and Technological Development (CNPq), grants nº 309201/2021-7 and 406949/2021-2.}

\markboth
{J.M.H Pinheiro \headeretal: The Impact of Feature Scaling In Machine Learning:
Effects on Regression and Classification Tasks}
{J.M.H Pinheiro \headeretal: The Impact of Feature Scaling In Machine Learning:
Effects on Regression and Classification Tasks}

\corresp{Corresponding author: João Manoel Herrera Pinheiro (e-mail: joao.manoel.pinheiro@usp.br).}

\begin{abstract}
This study addresses the lack of comprehensive evaluations of feature scaling by systematically assessing 12 techniques, including less common methods such as VAST and Pareto, in 14 machine learning algorithms and 16 datasets covering both classification and regression tasks. The impact of feature scaling was evaluated in terms of predictive performance (accuracy, MAE, MSE, R²) and computational efficiency (training time, inference time, memory usage). The results show that the ensemble methods (Random Forest, XGBoost, CatBoost, LightGBM) remain robust regardless of scaling, while models such as Logistic Regression, SVM, TabNet, and MLP are highly sensitive to the chosen scaler. By making all codes, results, and model parameters publicly available, this work provides reproducible, model-specific guidance for selecting scaling strategies in practical machine learning applications.
\end{abstract}

\titlepgskip=-21pt

\maketitle

\section{Introduction}
\label{sec:introduction}
Machine learning progress has been remarkable in several domains of knowledge engineering, notably driven by the rise of big data \cite{ZHOU2017350,10.1007/s10115-007-0114-2}, its applications in healthcare \cite{8474918,doi:10.1056/NEJMra2302038,doi:10.1056/NEJMp1606181,WANG20101519}, forecasting \cite{Ahmed30082010,https://doi.org/10.1111/joes.12429,agriculture13091671}, precision agriculture \cite{9311735}, wireless sensor networks \cite{6805162}, language tasks \cite{conneau2017deepconvolutionalnetworkstext} and many other domains \cite{8697857,10.1109/ICSC60394.2023.10441540,4783080,10.1016/j.neucom.2016.12.038,6296526,little2019machine}. All different applications compose the field of Machine Learning \cite{vaswani2023attentionneed}, which has become a major subarea of computer science and statistics due to its crucial role in the modern world \cite{Donoho02102017}. Although these methods hold immense potential for the advancement of predictive modeling, their improper application has introduced significant obstacles \cite{sculley2015hidden,10.1007/978-3-319-99740-7_1,inproceedingsKaufman}.

One such obstacle is the indiscriminate use of preprocessing techniques, particularly feature scaling \cite{garcia2015data}. Feature scaling is a mapping technique in a preprocessing stage by which the user tries to give all attributes the same weight \cite{han2011data,hastie}. In some applications, this data transformation can improve the performance of Machine Learning models \cite{10.3844/jcssp.2006.735.739}. 


Consequently, applying a scaling method without a careful evaluation of its suitability for the specific problem and model may not be advisable and could negatively impact results. This practice risks compromising the validity of claims regarding model performance and may increase the risk of overfitting, that is, when a model adapts too closely to noise or idiosyncrasies in the training data, thereby reducing its generalization capability\mbox{\cite{doi:10.1021/ci0342472}.}

Reproducibility is a critical problem in Machine Learning \cite{gundersen2023sourcesirreproducibilitymachinelearning,mcdermott2019reproducibilitymachinelearninghealth,semmelrock2023reproducibilitymachinelearningdrivenresearch,Haibe_Kains_2020}. It is often undermined by factors such as missing data or code, inconsistent standards, and sensitivity to training conditions \cite{https://doi.org/10.1002/aaai.70002}. Feature scaling, in particular, if not documented or applied correctly, can significantly affect model performance and hinder the replication of results. The absence of rigorous evaluation not only hampers reproducibility, but can also lead to the adoption of practices with poor generalizability across different datasets or domains. 

As Machine Learning methods continue to shape research by their use in a wide range of applications, it is essential to critically assess and justify each step of the modeling pipeline, including feature scaling, to ensure robust and replicable findings \cite{KAPOOR2023100804}.

The primary objective of this study is to evaluate the impact of different data scaling methods on the training process and performance metrics of various Machine Learning algorithms across multiple datasets. We employ 14 widely used Machine Learning models for tabular data, including Linear Regression, Logistic Regression, Support Vector Machines (SVMs), K-Nearest Neighbors (KNN), Multilayer Perceptron, Random Forest, TabNet, Naive Bayes, Classification and Regression Trees (CART), Gradient Boosting Trees, AdaBoost, LightGBM, CatBoost, and XGBoost. These models were evaluated using 12 different data scaling techniques, in addition to a baseline without scaling, across 16 datasets covering both classification and regression tasks. The selected models represent the state of the art in tabular data analysis, offering a favorable balance between predictive performance and computational efficiency, often outperforming deep learning techniques in this context \cite{SHWARTZZIV202284,10.5555/3540261.3541708,levin2023transferlearningdeeptabular,ye2025closerlookdeeplearning}.

This study contributes by being the first to address the lack of comprehensive evaluation in feature scaling and algorithm benchmarking:
\begin{itemize}
    \item Testing 12 feature scaling techniques, including less common methods such as VAST, Pareto, Logistic Sigmoid, and Hyperbolic Tangent.
    \item Benchmarking across 14 supervised learning algorithms, covering both regression and classification tasks.
    \item Employing 16 diverse datasets from different domains to ensure generalizability of the findings.
    \item Providing open-source code, experimental data, and parameters to ensure full reproducibility.
\end{itemize}

In Section \mbox{\ref{related:work}} we cover some related work in similar studies. In Section \mbox{\ref{related:back}} we explain more about each algorithm, each feature scaling technique and valuation metrics while Section \mbox{\ref{related:meto}}, the study diagram and how the models were trained. In Section \mbox{\ref{related:res}} we represent the final results of this study and some discussion. The limitations of our study are discussed in Section \mbox{\ref{related:lim}}. Lastly, in Section \mbox{\ref{sec:conclusion}} we give a final conclusion of the current experiments and future works.

\section{Related Work}\label{related:work}

For some Machine Learning models, feature scaling is extremely necessary, such as K-Nearest Neighbors\mbox{\cite{AKSOY2001563}}, Neural Networks\mbox{\cite{jayalakshmi2011statistical,6898836,mello2018machine}}, and SVMs\mbox{\cite{hsu2003practical,10.1007/978-981-10-2777-2_7}}.

Despite its fundamental role in Machine Learning pipelines, the impact of feature scaling remains an underexplored area in the literature. Most existing studies examine only a limited number of algorithms or datasets, and often provide minimal analysis of the specific effects of different scaling techniques\mbox{\cite{MAHARANA202291}} on each particular algorithm's performance. Studies that comprehensively evaluate various scaling methods across a broad range of models and datasets, such as the approach taken in this work, are scarce. In many Machine Learning cases, preprocessing is briefly mentioned, with scaling treated as a routine step rather than a variable worthy of in-depth investigation.

In\mbox{\cite{LI2011256}}, the authors compared 6 normalization methods applied only in an SVM classifier used to improve intrusion data. In the context of glaucoma detection based on a combination of texture and higher-order spectral features\mbox{\cite{5720314}}, the authors demonstrate that Z-score normalization, when paired with a Random Forest classifier, achieves superior performance compared to a Support Vector Machine.

In\mbox{\cite{Cao_Stojkovic_Obradovic_2016}}, the authors demonstrate how feature scaling methods can impact the final model performance; however, only 2 algorithms were used in a binary classification task. A more complete work\mbox{\cite{10.1093/bib/bbx153}} evaluates 12 learning algorithms and 6 different normalization techniques.

An in-depth study on the impact of data normalization on classification performance was presented in\mbox{\cite{SINGH2020105524}}, where the authors evaluated 14 normalization methods but employed only the K-Nearest Neighbor Classifier. Another study also used K-Nearest Neighbor Classifier and two SVMs\mbox{\cite{9214160}} with 7 scaling techniques, but only in 1 dataset.

In\mbox{\cite{technologies9030052}}, the authors evaluated 11 Machine Learning algorithms, across 6 different data scaling methods. However, they focused on only 1 dataset, the UCI - Heart Disease\mbox{\cite{misc_heart_disease_45}}.

In a study on a single dataset (diabetes diagnosis), researchers tested models including Random Forest, Naive Bayes, KNN, Logistic Regression, and SVM\mbox{\cite{9898687}}, but compared only 3 preprocessing scenarios.

A recent study\mbox{\cite{10681438}} evaluated 5 Machine Learning models, focusing solely on classification. However, some approaches normalized the entire dataset before splitting, introducing data leakage.

\begin{table}[!ht]
\caption{Comparative summary of related work in feature scaling.}
\label{table:work}
\setlength{\tabcolsep}{5.3pt}
\renewcommand{\arraystretch}{1.1}
\begin{tabularx}{\linewidth}{cccccX}
\toprule
\textbf{Paper} & \textbf{Year} & \textbf{Scalers} & \textbf{Algorithms} & \textbf{Datasets} & \multicolumn{1}{c}{\textbf{Gap}}                                                                                                  \\ \toprule
        \cite{LI2011256}       & 2011          & 6                   & 1                      & 1                    & \begin{tabular}[c]{@{}l@{}}Limited to \\ 1 algorithm \\and 1 dataset\end{tabular}                            \\ \hline
        \cite{5720314}      & 2011          & 2                   & 4                      & 1                    & \begin{tabular}[c]{@{}l@{}}Limited to \\ 2 scalers\\and 1 dataset\end{tabular}                              \\ \hline
        \cite{Cao_Stojkovic_Obradovic_2016}       & 2016          & 3                   & 2                      & 16                   & \begin{tabular}[c]{@{}l@{}}Only 2 algo-\\rithms and binary \\ classification\end{tabular}                     \\ \hline
        \cite{10.1093/bib/bbx153}      & 2017          & 2                   & 12                     & 1                    & \begin{tabular}[c]{@{}l@{}}Limited to\\ 2 scalers\\and 1 dataset\end{tabular}                                \\ \hline
         \cite{SINGH2020105524}      & 2019          & 14                  & 1                      & 21                   & \begin{tabular}[c]{@{}l@{}}Focused only\\ on 1 algorithm\end{tabular}                                         \\ \hline
         \cite{9214160}      & 2020          & 6                   & 3                      & 1                    & \begin{tabular}[c]{@{}l@{}}Limited to\\3 algorithms and\\ only 1 dataset\end{tabular}                       \\ \hline
        \cite{technologies9030052}       & 2021          & 6                   & 11                     & 1                    & \begin{tabular}[c]{@{}l@{}}Only 1 dataset\end{tabular}                                        \\ \hline
         \cite{9898687}      & 2022          & 3                   & 5                      & 1                    & \begin{tabular}[c]{@{}l@{}}Restricted to\\ 1 dataset\\and 3 scalers\end{tabular}                             \\ \hline
         \cite{10681438}      & 2024          & 2                   & 5                      & 1                    & \begin{tabular}[c]{@{}l@{}}Only 2 algo-\\rithms, 1 dataset,\\and scaler applied\\before splitting\end{tabular} \\ \toprule
\end{tabularx}
\end{table}

Compared to prior studies (see Table\mbox{~\ref{table:work}}) our work offers a broader and more systematic evaluation of feature scaling. We benchmark 12 techniques across 14 algorithms and 16 datasets. Unlike earlier research, we ensure reproducibility by releasing open-source code, parameters, and data.

\section{Background}\label{related:back}
\subsection{Feature Scaling Techniques}
Several feature scaling techniques were investigated. For a subset of these, we leveraged the built-in implementations available in the \texttt{scikit-learn} library. However, other specialized or less common scaling methods required custom implementation, which we developed as classes within our Python experimental framework.

We selected 12 feature scaling techniques to balance breadth, diversity, and experimental feasibility across the 14 algorithms and 16 datasets considered. This set includes widely adopted methods such as Min–Max, Standardization, and Robust Scaling, which represent standard practice in Machine Learning, alongside less commonly benchmarked but theoretically relevant approaches such as VAST, Pareto, Logistic Sigmoid, and Hyperbolic Tangent, which offer distinctive transformations with potential impact on model performance. While additional scalers exist, including too many would make the experiments intractable and the results harder to interpret. By focusing on these 12 methods, we provide a comprehensive yet manageable benchmark that captures both established practices and underexplored alternatives, ensuring the study’s value as a practical reference for future work.

\vspace{0.2cm}
\subsubsection{Min-Max Normalization (MM)}
Min-Max normalization scales the data to a fixed range, typically [0, 1] \cite{han2011data,JAIN20052270}. The transformation is given by:
\begin{equation}
X_{\text{norm}} = \frac{X - X_{\text{min}}}{X_{\text{max}} - X_{\text{min}}},
\end{equation}
where $X$ is the original value, $X_{\text{min}}$ and $X_{\text{max}}$ are the minimum and maximum values in the dataset, $X_{\text{norm}}$ is the normalized value.

\vspace{0.2cm}
\subsubsection{Max Normalization (MA)}
Max normalization scales the data by dividing each feature by its maximum absolute value \cite{han2011data,10.1007/978-3-031-42536-3_33}:
\begin{equation}
X_{\text{norm}} = \frac{X}{max(~|X|~)}
\end{equation}
This method is advantageous when the data consists of strictly non-negative values.

\vspace{0.2cm}
\subsubsection{Z-score Normalization (ZSN)}
Z-score normalization (also known as Standardization) transforms data to have a mean of 0 and a unit variance \cite{han2011data,10.1093/bioinformatics/bti124}. The formula is given by:

\begin{equation}
X_{\text{norm}} = \frac{X - \mu}{\sigma},
\end{equation}
where $X$ is the original feature value, $\mu$ is the mean of the feature, $\sigma$ is the standard deviation of the feature, and $X_{\text{norm}}$ is the scaled value.

\vspace{0.2cm}
\subsubsection{Variable Stability Scaling (VAST)}
Variable stability scaling adjusts the data based on the stability of each feature. It is particularly useful for high-dimensional datasets and can be seen as a variation of standardization that incorporates the Coefficient of Variation (CV), $\frac{1}{CV}=\frac{\mu}{\sigma}$, as a scaling factor \cite{KEUN2003265,van2006centering}:
\begin{equation}
X_{\text{norm}} = \frac{(X - \mu)}{\sigma} \frac{\mu}{\sigma}
\end{equation}

\vspace{0.2cm}
\subsubsection{Pareto Scaling (PS)}
Pareto scaling is a normalization technique in which each feature is centered, by subtracting the mean, and then divided by the square root of its standard deviation \cite{NODA2008216}. It's similar to Z-score normalization, but instead of dividing by the full standard deviation, it uses its square root as the scaling factor. Pareto scaling is particularly useful when the goal is to preserve relative differences between features while reducing the impact of large variances \cite{eriksson1999introduction,kubinyi20063d}.

\begin{equation}
X_{\text{norm}} = \frac{X - \mu}{\sqrt{\sigma}}
\end{equation}

\vspace{0.2cm}
\subsubsection{Mean Centered (MC)}
Mean centering subtracts the mean of each feature from the data. This method is often used as a preprocessing step in Principal Component Analysis (PCA) \cite{kim2024pcasvdcenteringdata}. 
\begin{equation}
X_{\text{norm}} = X - \mu
\end{equation}

\vspace{0.2cm}
\subsubsection{Robust Scaler (RS)}
The robust scaler uses the median and interquartile range (IQR) to scale the data:
\[
X_{\text{norm}} = \frac{X - X_{\text{median}}}{\text{IQR}}
\]
This method is robust to outliers \cite{10.1007/978-3-031-75823-2_18,DEAMORIM2023109924}.

\vspace{0.2cm}
\subsubsection{Quantile Transformation (QT)}
Quantile transformation maps the data to a uniform or normal distribution. It is useful for non-linear data \cite{Bolstad2003,Kelmansky2013}. 

\vspace{0.2cm}
\subsubsection{Decimal Scaling Normalization (DS)}
Decimal scaling performs normalization by adjusting the decimal point of the attribute values, thereby rescaling them to fit within the range (–1, 1), not including the endpoints \cite{garcia2015data,han2011data}:
\begin{equation}
X_{\text{norm}} = \frac{X}{10^j}
\end{equation}
where $j$ is the smallest integer such that $max(~|X_{norm}|~)<1$

\vspace{0.2cm}
\subsubsection{Tanh Transformation (TT)}
A variant of tanh normalization is used, in which the Hampel estimators are replaced by the mean and standard deviation of each feature \cite{10.1109/TPAMI.2005.57}:
\begin{equation}
X_{\text{norm}} = \frac{1}{2} \left\{ \tanh\left(0.01 ~  \frac{X - \mu}{\sigma} \right) + 1 \right\}
\end{equation}

\vspace{0.2cm}
\subsubsection{Logistic Sigmoid Transformation (LS)}
The logistic sigmoid-based transformation applies the sigmoid function to the data \cite{10.5555/1457541,priddy2005artificial}:
\begin{equation}
X_{\text{norm}}  = \frac{1}{1 + e^{-q}}, \quad \text{where} \quad q = \frac{X - \mu}{\sigma}
\end{equation}

\vspace{0.2cm}
\subsubsection{Hyperbolic Tangent Transformation (HT)}
The hyperbolic tangent transformation is similar to the tanh transformation but is applied differently in certain contexts \cite{10.5555/1457541,priddy2005artificial}:
\begin{equation}
X_{\text{norm}}  = \frac{1- e^{-q}}{1 + e^{-q}}, \quad \text{where} \quad q = \frac{X - \mu}{\sigma}
\end{equation}

\subsection{Machine Learning Algorithms}

\vspace{0.2cm}

The following models were used:

\begin{itemize}
    \item \textbf{Logistic Regression (LR):} A simple and effective statistical model for binary classification that estimates class probabilities using the logistic function \cite{hosmer2013applied}.
    
    \item \textbf{Linear Regression:} A fundamental model that fits a linear relationship between input characteristics and a continuous target variable \cite{hastie,james2013introduction,https://doi.org/10.1002/wics.1198}

    \item \textbf{Support Vector Machine (SVM) \& Support Vector Regressor (SVR):} This model finds the hyperplane that maximizes the margin between classes and for regression finds a function within a tolerance margin using the radial basis function (RBF) kernel\mbox{\cite{cortes1995support,Ben-Hur2010,drucker1997support}}.

    \item \textbf{Multilayer Perceptron (MLP):} A feedforward neural network with one or more hidden layers trained using backpropagation, for classification and regression \cite{lecun2015deep,he2015delvingdeeprectifierssurpassing,kingma2017adammethodstochasticoptimization}.

    \item \textbf{Random Forest (RF):} An ensemble of decision trees built using bootstrap aggregation (bagging), improving robustness, and reducing overfitting, works for classification and regression \cite{breiman2001random,10.1007/978-3-030-03146-6_86,Qi2012}.

    \item \textbf{Naive Bayes (NB):} Based on Bayes' Theorem with the assumption of conditional independence between features, only works for classification \cite{rish2001empirical}.

    \item \textbf{Classification and Regression Trees (CART):} A recursive partitioning algorithm that builds binary trees for decision-making, works for classification and regression tasks \cite{breiman1984classification,SINGHKUSHWAH20223571}.

    \item \textbf{LightGBM (LGBM):} A gradient boosting framework that uses histogram-based learning and leaf-wise tree growth, works for classification and regression \cite{ke2017lightgbm}.

    \item \textbf{AdaBoost (Ada):} A boosting algorithm that iteratively focuses on misclassified examples to improve the accuracy of the model, works for classification and regression \cite{freund1997decision,Schapire2003}.

    \item \textbf{CatBoost:} A gradient boosting model with native support for categorical features, works for regression and classification \cite{dorogush2018catboost}.

    \item \textbf{XGBoost:} An efficient implementation of gradient boosting with regularization and optimized parallel computing, works for regression and classification \cite{chen2016xgboost}.

    \item \textbf{K-Nearest Neighbors (KNN):} An instance-based learning algorithm that classifies samples based on the most frequent label among their nearest neighbors, also has a regression variant that predicts the target by averaging the outputs of the $k$ nearest neighbors in the feature space \cite{cover1967nearest,7898482,10.1145/2990508}.

    \item \textbf{Attentive Interpretable Tabular Learning (TabNet):} A deep learning model for tabular data that employs sequential attention to select relevant features at each decision step, works for classification and regression \cite{arik2020tabnetattentiveinterpretabletabular}.
\end{itemize}

\vspace{0.2cm}

\subsection{Metrics}
\subsubsection{Classification Metrics}
\vspace{0.2cm}

\begin{itemize}
    \item \textbf{Accuracy:} one of the most widely used metrics to evaluate classification tasks. Measures the proportion of correctly predicted instances relative to the total number of predictions: 
\begin{equation}
\text{Accuracy} = \frac{TP + TN}{TP + TN + FP + FN},
\label{accuracy_equation}
\end{equation}
where $TP$, $TN$, $FP$, and $FN$ represent true positives, true negatives, false positives, and false negatives, respectively.

Despite its popularity, accuracy can be misleading when dealing with imbalanced datasets, as highlighted by \cite{powers2011evaluation}. However, given the characteristics of our datasets, comprising both binary and multiclass classification problems, we chose to include accuracy in our analysis, acknowledging its limitations in the presence of class imbalance.
\end{itemize}

\vspace{0.2cm}
\subsubsection{Regression Metrics} 
\vspace{0.2cm}
\begin{itemize}
    \item \textbf{Mean Absolute Error (MAE):} measures the average absolute difference between predicted and actual values, offering an intuitive sense of the magnitude of the error.

\begin{equation}
\text{MAE} = \frac{1}{n} \sum_{i=1}^{n} |y_i - \hat{y}_i|
\end{equation}

\vspace{0.2cm}
\item \textbf{Mean Squared Error (MSE):} calculates the average squared differences between actual and predicted values, penalizing larger errors more heavily:

\begin{equation}
\text{MSE} = \frac{1}{n} \sum_{i=1}^{n} (y_i - \hat{y}_i)^2
\end{equation}

\vspace{0.2cm}
\item \textbf{Coefficient of Determination ($R^2$):} indicates the proportion of variance in the dependent variable that is predictable from the independent variables. A higher value $R^2$ indicates a better fit of the model to the data \cite{10.1093/biomet/78.3.691}.
\end{itemize}
These regression metrics are standard for evaluating continuous output predictions \cite{chicco2021coefficient}.

\subsubsection{Computational Metrics}
To complement the evaluation of predictive performance, we also assess:
\begin{itemize}
    \item \textbf{Memory Usage}: Measure the increase in memory usage during dataset loading and the feature scaling step.
    \item \textbf{Training Time}: The time taken to train each model on a given dataset.
    \item \textbf{Inference Time}: The time required for the trained model to make predictions on unseen data.
\end{itemize}
These metrics are essential when evaluating models for real-time or resource-constrained environments.

\section{Methodology}\label{related:meto}
Our primary focus in this study is to ensure reproducibility. To that end, we use a well-known dataset for classification and regression tasks, sourced from the University of California, Irvine (UCI) Machine Learning Repository, due to its well-recognized and diverse collection of real-world datasets with standardized formats, that can easily be used for benchmarking and comparison between the different models chosen in this work.
 
\subsection{Dataset}

Tables \ref{table:data:class} and \ref{table:data:reg} provide detailed information about the datasets used in this study, including the number of features, instances, and classes. All features are numeric, represented as either \texttt{int64} or \texttt{float64} types. The classification tasks are either binary or multi-class.

\begin{table}[!ht]
\caption{Datasets used for classification:  Breast Cancer Wisconsin (Diagnostic) \cite{misc_breast_cancer_wisconsin_(diagnostic)_17}, Dry Bean Dataset \cite{misc_dry_bean_dataset_602}, Glass Identification \cite{misc_glass_identification_42}, Heart Disease \cite{misc_heart_disease_45}, Iris \cite{misc_iris_53}, Letter Recognition \cite{misc_letter_recognition_59}, MAGIC Gamma Telescope \cite{misc_magic_gamma_telescope_159}, Rice (Cammeo and Osmancik) \cite{misc_rice_(cammeo_and_osmancik)_545}, and Wine \cite{misc_wine_109}.}
\label{table:data:class}
\centering
\begin{tabular}{cccc}
\toprule
Dataset & Instances & Features & Classes\\
\toprule
Breast Cancer Wisconsin (Diagnostic) & 569 & 30 & 2 \\
Dry Bean Dataset & 13611 & 16 & 7 \\
Glass Identification & 214 & 9 & 6\\
Heart Disease & 303 & 13 & 2 \\
Iris & 150 & 4 & 3 \\
Letter Recognition & 20000 & 16 & 26 \\
MAGIC Gamma Telescope & 19020 & 10 & 2 \\
Rice (Cammeo and Osmancik) & 3810 & 7 & 2 \\
Wine & 178 & 13 & 3 \\
\toprule
\end{tabular}
\end{table}

\begin{table}[!ht]
\caption{Datasets used for regression:  Air Quality \cite{air_quality_360}, Abalone \cite{abalone_1}, Appliances Energy Prediction \cite{appliances_energy_prediction_374}, Concrete Compressive Strength \cite{concrete_compressive_strength_165}, Forest Fires \cite{forest_fires_162}, Real Estate Valuation \cite{real_estate_valuation_477}, and Wine Quality \cite{wine_quality_186}.}
\label{table:data:reg}
\centering
\begin{tabular}{ccc}
\toprule
Dataset & Instances & Features \\\toprule
Abalone & 4177 & 8  \\
Air Quality & 9358 & 15  \\
Appliances Energy Prediction & 19735 & 28  \\
Concrete Compressive Strength & 1030 & 8  \\
Forest Fires & 517 & 12  \\
Real Estate Valuation & 414 & 6  \\
Wine Quality & 4898 & 11  \\
\toprule
\end{tabular}
\end{table}

\subsection{Train and Test set}
To preserve the integrity of our analysis and avoid data leakage, we split the dataset into training and test sets prior to applying any preprocessing steps, such as feature scaling. 

Data leakage occurs when information from the test set inadvertently influences the training process, often leading to overly optimistic performance estimates. For instance, performing oversampling or other transformations \emph{before} splitting the data can introduce overlap between the training and test sets, thereby compromising their independence. By strictly separating the data prior to any preprocessing, we maintain a clear boundary between the two sets, ensuring a robust and unbiased evaluation of the model performance\mbox{\cite{KAPOOR2023100804}}.

Following standard practice in machine learning\mbox{\cite{raschka2020modelevaluationmodelselection}},\linebreak\mbox{\cite{info:doi/10.2196/49023}}, 70\% of the data was allocated to the training set and 30\% to the test set. There is no universal optimal train–test split in Machine Learning\mbox{\cite{https://doi.org/10.1002/sam.11583}}. The choice depends on dataset size, the bias–variance trade-off, and computational constraints. In this study, we adopted a 70/30 split as a practical balance that allows sufficient data for model training while still allowing reliable performance evaluation. To ensure reproducibility, random seeds were fixed in all experiments.

\subsection{Experiment Workflow}
The objective of this subsection is to document every step of the experimentation process to ensure full transparency and reproducibility. All datasets used in this work are publicly available and can be downloaded along with their respective train/test splits. All Machine Learning algorithms were applied using their default hyperparameters, as implemented in \texttt{scikit-learn} or the corresponding official libraries. For algorithms that support the \texttt{random\_state} parameter, a fixed seed was used to ensure consistent results across runs. During each training session, a configuration file is generated to record the parameters used by each model, enabling complete traceability of the experimental setup. 

The experiment begins with the import and cleaning of each dataset. Categorical target variables are encoded numerically, and column names are standardized using regular expressions. After that, each dataset is partitioned into training and testing subsets, which are saved both as \texttt{.csv} files and as Python dictionaries. Subsequently, for every dataset and Machine Learning model combination, various scaling techniques are applied. Each model is trained on the training set and evaluated on the test set, with performance metrics computed accordingly. In addition to the results, the model configuration and metadata - including training and inference times - are stored to ensure reproducibility and to facilitate further analysis.
\begin{algorithm}[H]
    \caption{Experiment Workflow}
    \label{alg:simplified_ml_pipeline}
    \begin{algorithmic}[1]
        \Require $D$: Training Data
        \Ensure Validation metrics, training time, inference time, memory usage, and model parameters.

        \State \textbf{Begin}
        \State Import necessary libraries and load dataset $D$.
        \State Perform cleaning and ETL (Extract, Transform, Load) on $D$.
        \State Split $D$ into $D_{train}$ (training set) and $D_{test}$ (testing set).

        \For{each available dataset}
            \For{each Machine Learning Model $M$}
                \State Fit scaler on $D_{train}$ and apply transform to $D_{train}$ and $D_{test}$.
                \State Train Model $M$ using $D_{train}$.
                \State Evaluate $M$ on $D_{test}$.
                \State Save validation results, including accuracy, training time, inference time, memory usage, and model parameters.
            \EndFor
        \EndFor
        \State \textbf{End}
    \end{algorithmic}
\end{algorithm}
\subsection{Python Script Descriptions}

The following scripts were developed to automate and manage the experimental pipeline:

\begin{itemize}
    \item \texttt{import\_dataset.py}: Imports datasets from the UCI repository and maps the appropriate target variable for each dataset.

    \item \texttt{etl\_cleaning.py}: Convert categorical and numerical variables as needed and cleans column names using regular expressions.

    \item \texttt{train\_test\_split.py}: Splits the datasets into training and testing sets and saves them in both \texttt{.csv} and dictionary formats.

    \item \texttt{train\_results.py}: Train each Machine Learning model on every dataset using different scaling techniques. Calculate validation metrics and save both performance results and model configuration.

    \item \texttt{main.py}: Serves as the main execution script that orchestrates all stages of the experiment.
\end{itemize}

\begin{figure}[htbp]
    \centering
    \includegraphics[width=0.5\textwidth]{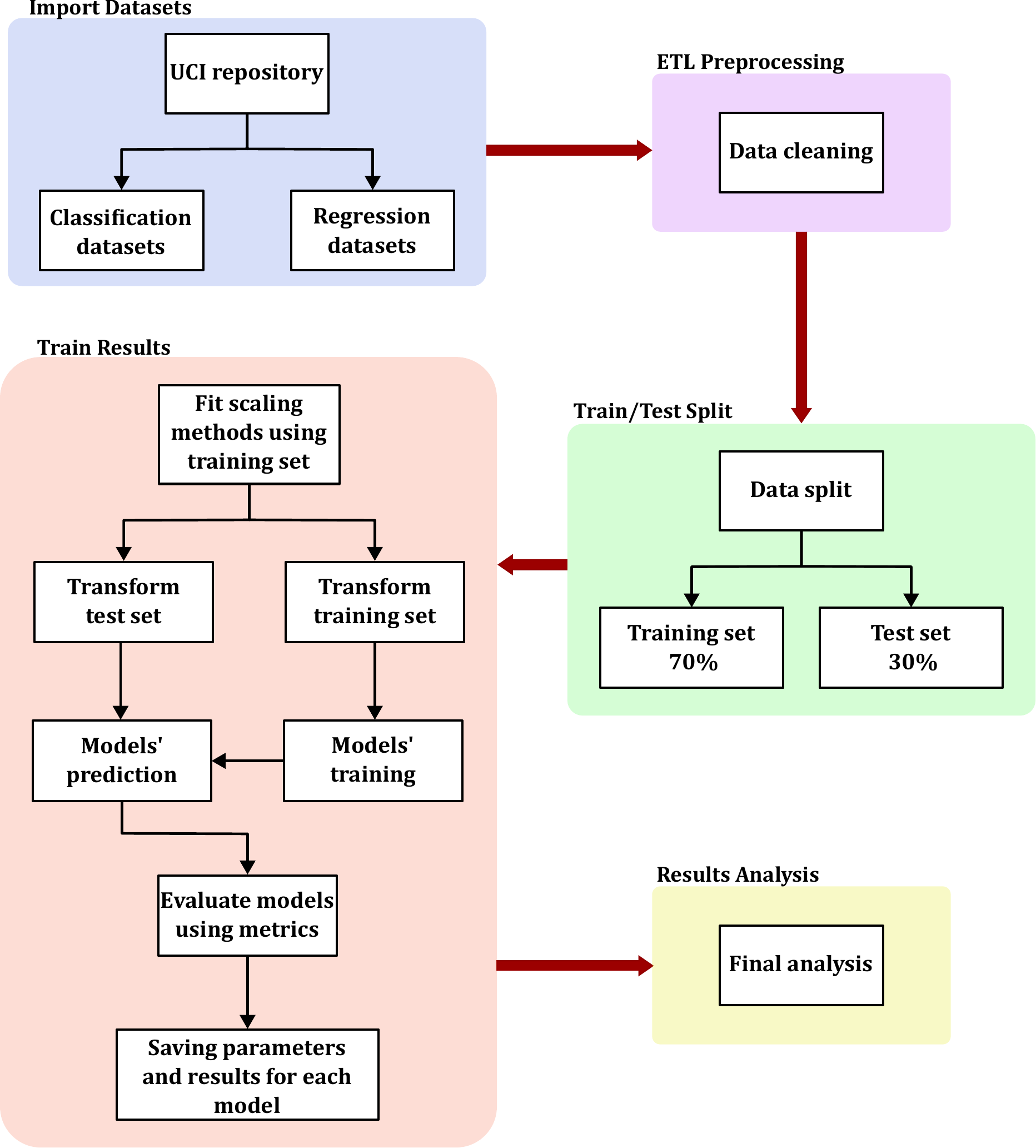}
    \caption{Experimental design.}
    \label{fig:flow}
\end{figure}
\subsection{Source Code of this Experiments}
To ensure full transparency and reproducibility, we provide the source code, complete experimental results, statistical analyses, detailed model parameters, and all accompanying plots and figures. They are publicly available in \mbox{\href{https://github.com/joaomh/article-impact-feature-scaling-classification}{GitHub}}.

The experiments were conducted on a system running a 64-bit Linux distribution, equipped with an AMD Ryzen™ 9 7900 processor (capable of boosting to 5.4 GHz) and 64 GB of RAM.
\section{Results \& Discussion}\label{related:res}
This section presents the empirical results of our extensive experiments. We selected 5 representative models and 3 scaling techniques, along with a baseline without scaling, as this subset already allows us to draw meaningful conclusions and observe distinctions among models. The complete results are available in the Appendix \mbox{\ref{appendix:A}}.

\subsection{Impact on Validation Metrics}

We used the Wilcoxon signed-rank test\mbox{\cite{wilcoxon1945,10.5555/1248547.1248548}} to compare the performance of the algorithms on scaled and unscaled datasets and the Friedman test\mbox{\cite{10.1214/aoms/1177731944}} to determine the statistical significance of the data scaling results. We adopted a significance threshold of 0.01 to reduce the likelihood of Type I errors (false positives) when comparing models with and without feature scaling. Given the multiple datasets and models involved, a stricter alpha level provides greater confidence that any observed differences in performance are truly meaningful. The complete results of these statistical tests are in Appendix \mbox{\ref{appendix:A}}.

\begin{figure*}[htbp]
    \centering
    \includegraphics[width=1.0\textwidth]{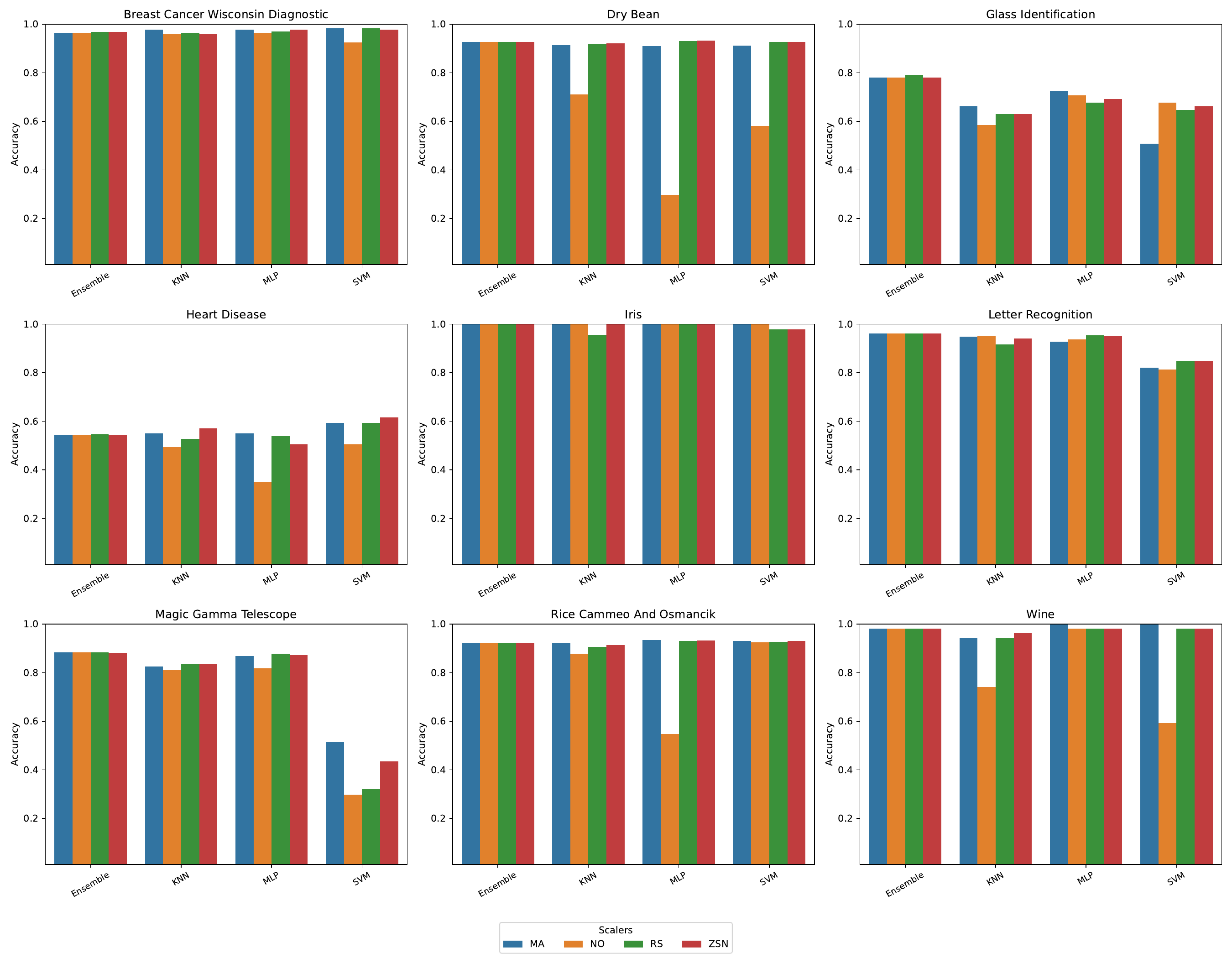}
    \caption{Accuracy results for the 9 datasets.  
    The ensemble models are: RF, LGBM, CatBoost, and XGBoost.}
    \label{fig:all_datasets_accuracy}
\end{figure*}

As anticipated, one of the key findings of our classification experiments is the differential impact of feature scaling on model performance. Ensemble methods, including Random Forest and the gradient boosting family (LightGBM, CatBoost, XGBoost), demonstrated strong robustness by consistently achieving high validation performance regardless of the preprocessing strategy or dataset. The Naive Bayes model showed similar scaling resistance, although its overall accuracy was not competitive with these top-tier ensembles. This observation is statistically supported by the Wilcoxon signed-rank test comparing scaled versus unscaled features. These results are confirmed by an analysis of Figure \mbox{\ref{fig:all_datasets_accuracy}} and Tables \mbox{\ref{table:stati:resume}}, \mbox{\ref{tab:accuracy_stests}}, and \mbox{\ref{tab:accuracy_big_table}}.

In contrast, the performance of LR, SVM, KNN, TabNet, and MLP were highly dependent on the choice of scaler, revealing their pronounced sensitivity to data preprocessing (see Figure \mbox{\ref{fig:all_datasets_accuracy}} and Table \mbox{\ref{tab:accuracy_big_table}}). This observation is statistically supported by the Wilcoxon signed-rank test comparing scaled versus unscaled features, as shown in Tables \mbox{\ref{table:stati:resume}}, and \mbox{\ref{tab:accuracy_stests}}.

 Similarly, the Friedman test across all scaling methods confirmed significant performance variation for LR, SVM, MLP, and KNN, while RF, NB, CART, AdaBoost, CatBoost, LGBM, XGBoost, and TabNet showed no significant differences. These results quantitatively support the observed trends, highlighting which models are inherently scaling-sensitive versus scaling-robust.

\begin{figure*}[htbp]
    \centering
    \includegraphics[width=1.0\textwidth]{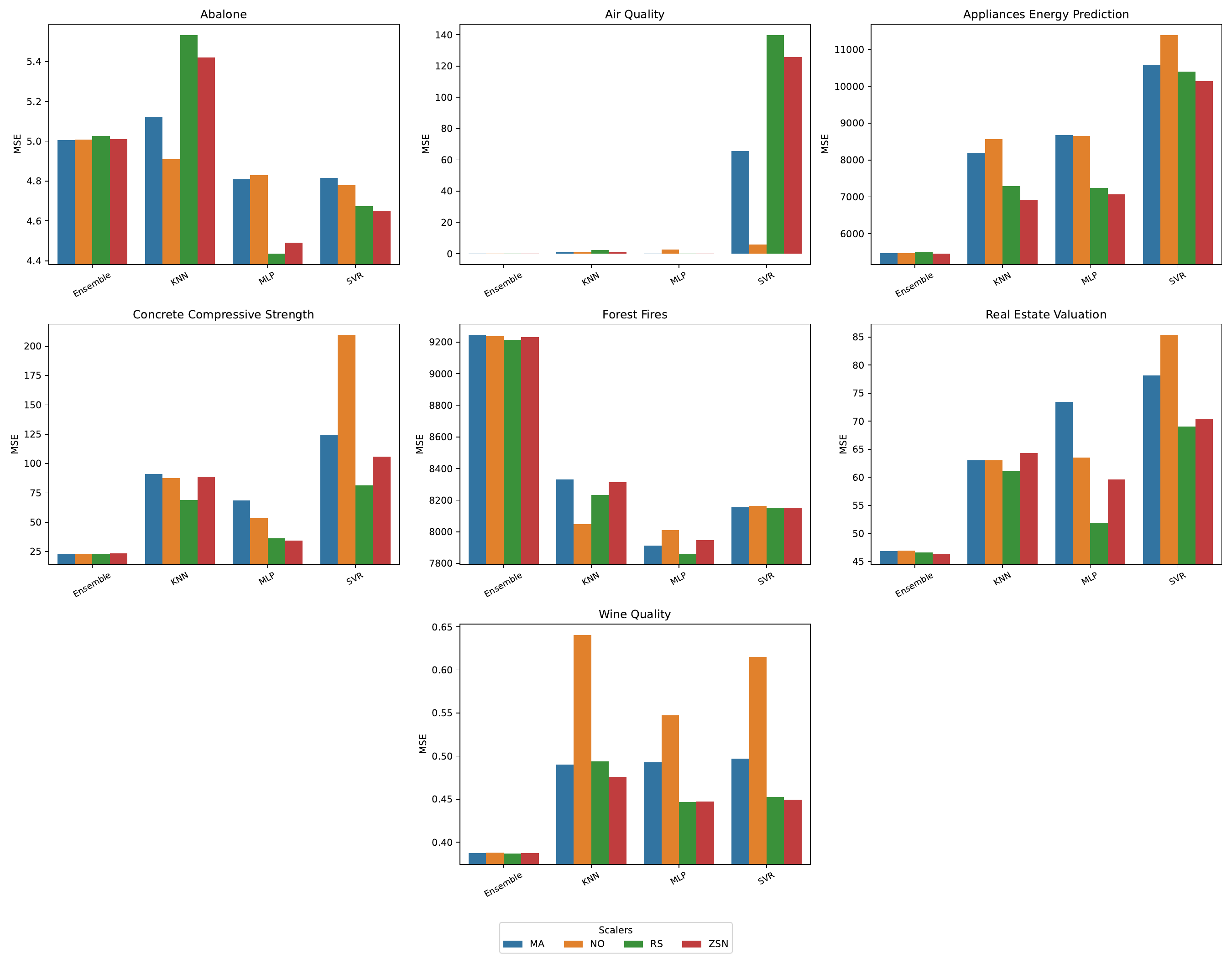}
    \caption{MSE results for the 7 datasets.  
    The ensemble models are: RF, LGBM, CatBoost, and XGBoost.}
    \label{fig:all_datasets_mse}
\end{figure*}

In regression tasks, a similar pattern of scaling sensitivity was observed when employing the regression counterparts of these classification models (see Figures \mbox{\ref{fig:all_datasets_mse}}, \mbox{\ref{fig:all_datasets_r2}}, and \mbox{\ref{fig:all_datasets_mae}}), with the results from the Wilcoxon signed-rank and Friedman tests shown in Tables \mbox{\ref{table:stati:resume}}, \mbox{\ref{tab:r2_stests}}, \mbox{\ref{tab:mse_stests}}, and \mbox{\ref{tab:mae_stests}}. Notably, for KNN in regression, the Wilcoxon test did not indicate statistically significant differences across scaling methods. This suggests that the underlying mathematical principles governing these model families lead to consistent behavior regarding data scaling, whether applied to classification or regression problems. The complete results are presented in Tables \mbox{\ref{tab:r2_big_table}}, \mbox{\ref{tab:mse_big_table}} and \mbox{\ref{tab:mae_big_table}}.

{\footnotesize
\setlength{\tabcolsep}{2pt} 
\begin{table}[!ht]
\caption{Wilcoxon test results (scaled vs. unscaled features) and Friedman test results (across all scaling methods) for classification tasks and regression tasks.}
\label{table:stati:resume}
\centering
\begin{tabular}{ccccc}
\toprule
Model & Wilcoxon $p$ & Wilcoxon sig. & Friedman $p$ & Friedman sig. \\
\toprule
SVM  & 0.0001 & Yes & 0 & Yes \\
SVR & 0 & Yes & 0.0007 & Yes \\
KNN & 0-0.6811 & Yes/No & 0.0393-0.0040& Yes/No \\
MLP & 0-0.0077 & Yes & 0-0.0006 & Yes \\
Ensemble & 0.0252-0.9710 & No & 0.1369-1.0 & No \\
\toprule
\end{tabular}
\end{table}}

In general, our findings affirm the superior performance of the ensemble methods, Random Forest, LightGBM, CatBoost, and XGBoost, which is consistent with their established reputation as state-of-the-art models for tabular data. Their ability to achieve high accuracy regardless of the scaling technique applied explains why feature scaling is often considered an optional preprocessing step for these particular models in many Machine Learning projects. This practical consideration, combined with their predictive power, underscores their utility in a wide range of applications.

\subsection{Impact on Training and Inference Times}

The application of different scaling techniques had a variable impact on inference times across the evaluated models. Notably, Classification and Regression Trees based models exhibited exceptionally robust behavior, remaining unaffected across all scaling methods and datasets. While certain Machine Learning algorithms, such as K-Nearest Neighbors (KNN), Support Vector Machine (SVM), and Support Vector Regressor (SVR) showed more evident sensitivity to the choice of scaling technique, as illustrated in Figures \mbox{\ref{fig:all_datasets_class_inf}} and \mbox{\ref{fig:all_datasets_reg_inf}}, this was not the norm. As shown in Tables \ref{tab:inf_class_big_table} and \ref{tab:inf_reg_big_table}, for the majority of models, the preprocessing step introduced only a small, non-uniform computational overhead, which may become more significant in the context of large-scale datasets or time-sensitive applications requiring real-time inference.

For training time, the effects of scaling largely mirrored those seen in the validation accuracy results, as shown in Figures \mbox{\ref{fig:all_datasets_class_train}} and \mbox{\ref{fig:all_datasets_ref_train}}. Certain models, notably tree-based ensembles, did not derive a significant speed benefit from feature scaling, while others were more sensitive. The complete results are available in Tables \mbox{\ref{tab:train_class_big_table}} and \mbox{\ref{tab:train_reg_big_table}}.

\subsection{Results of Memory Usage (kB)}
The analysis of Figure \mbox{\ref{fig:all_datasets_mem}} reveals a clear distinction in memory consumption between scaling techniques. As expected, applying no scaling resulted in zero additional memory consumption, beyond what is required to load the data. Among the actual scaling methods, the RobustScaler, StandardScaler, Tanh Transformer, and Hyperbolic Tangent were found to be the most memory intensive. In contrast, the MaxAbsScaler, MinMaxScaler, and Decimal Scaler consistently registered the lowest memory usage, as detailed in Table \ref{tab:memory_big_table}.

\section{Limitation}\label{related:lim}
While this study provides a broad empirical analysis of feature scaling across various models and datasets, certain limitations should be acknowledged, which also open avenues for future research.

\begin{itemize}
    \item \textbf{Hyperparameter Optimization:} The Machine Learning models analyzed in this study were evaluated using their default hyperparameters, as outlined in the methodology. A comprehensive hyperparameter tuning process for each model, scaler and dataset combination was beyond the current scope; however, such optimization could potentially uncover different optimal pairings or further improve model performance.

    \item \textbf{Scope and Diversity of Datasets:} Although 16 datasets were used for both classification and regression tasks, the findings could be further enhanced by incorporating an even wider array of datasets, particularly those with very high dimensionality, different types of underlying data distributions, or from more specialized domains.

    \item \textbf{Evaluation Metrics for Classification:} The primary metric for classification tasks was accuracy. Although acknowledged as potentially misleading for imbalanced datasets, future work could incorporate a broader suite of metrics, such as F1 score, precision, recall AUC, or balanced precision, to provide a more nuanced understanding of performance, especially on datasets with skewed class distributions.
\item \textbf{K-Fold Cross-Validation:} 
Although cross-validation is a widely recommended practice to obtain more reliable performance estimates, in this study, it was not employed due to the large number of datasets, models, and scaling configurations evaluated. Performing k-fold cross-validation in this experimental setup would substantially increase the computational cost, making the analysis impractical within the available resources. Instead, we adopted a hold-out validation strategy, which allowed for a consistent and feasible comparison across all scenarios. Future work could incorporate cross-validation or repeated runs to further validate the robustness of the findings.

    \item \textbf{Dataset Size and Synthetic Data:} For some of the smaller datasets utilized, the exploration of techniques such as synthetic data generation or data augmentation was not performed. Such methods could potentially improve the robustness and performance of certain models, representing a promising direction for further research.

    \item \textbf{Focus on Default Algorithm Implementations:} The study relied on standard implementations of algorithms mainly from well-known libraries. Investigating variations or more recent advancements within these algorithm families could offer additional insight.
\end{itemize}

These limitations are common in empirical studies of this nature and primarily highlight areas where this already extensive work could be expanded in the future.
\section{Conclusion}
\label{sec:conclusion}
This comprehensive empirical study investigated the impact of 12 feature scaling techniques on 14 Machine Learning algorithms in 16 classification and regression datasets. Key findings reaffirmed the robustness of ensemble methods (e.g., Random Forest, gradient boosting family), which, along with Naive Bayes, largely maintained high performance irrespective of scaling. This offers efficiency gains by potentially avoiding preprocessing overhead. In stark contrast, models such as Logistic Regression, SVMs, MLPs, K-Nearest Neighbor, and TabNet demonstrated high sensitivity, with their performance critically dependent on scaler choice. Computational analysis also indicated that scaling choices can influence training/inference times and memory usage, with certain scalers being notably more resource-intensive.

This study contributes with one of the first systematic evaluations of such an extensive array of models, some less common models like TabNet, and scaling techniques — including transformations, e.g Tanh (TT) and Hyperbolic Tangent (HT), which are less commonly benchmarked as general-purpose scalers —, all within a unified Python framework. This is particularly relevant given that feature scaling is often applied in the literature without a clear rationale, sometimes incorrectly before data splitting, which can lead to data leakage, or without verifying algorithm-specific benefits. By providing broad empirical evidence, our work offers clear guidance on how to mitigate these common issues, promoting informed scaling selection and more rigorous experimental design.

Future research could extend these insights by exploring extensive hyperparameter optimization, incorporating more diverse datasets, and utilizing a broader suite of evaluation metrics. Nevertheless, this study significantly contributes to a deeper, practical understanding of feature scaling's role in Machine Learning.

\ifCLASSOPTIONcaptionsoff
  \newpage
\fi



%

\bibliographystyle{IEEEtran}
\bibliography{ref}

%

\begin{IEEEbiography}[{\includegraphics[width=1in,height=1.25in,clip,keepaspectratio]{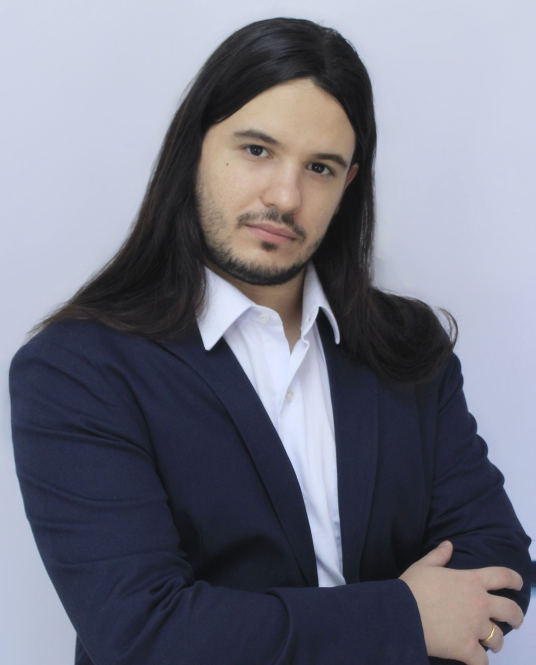}}]{João Manoel Herrera Pinheiro}
Received the B.Sc. degree in Mechatronics Engineering. Currently pursuing an M.Sc. degree at the University of São Paulo, with a focus on Computer Vision and Machine Learning. Currently enrolled in two specialization programs: Didactic-Pedagogical Processes for Distance Learning at UNIVESP and Software Engineering at USP. Serves as a reviewer for international journals such as Nature Scientific Data, Artificial Intelligence (IBERAMIA) and the Journal of the Brazilian Computer Society.
\end{IEEEbiography}

\begin{IEEEbiography}[{\includegraphics[width=1in,height=1.25in,clip]{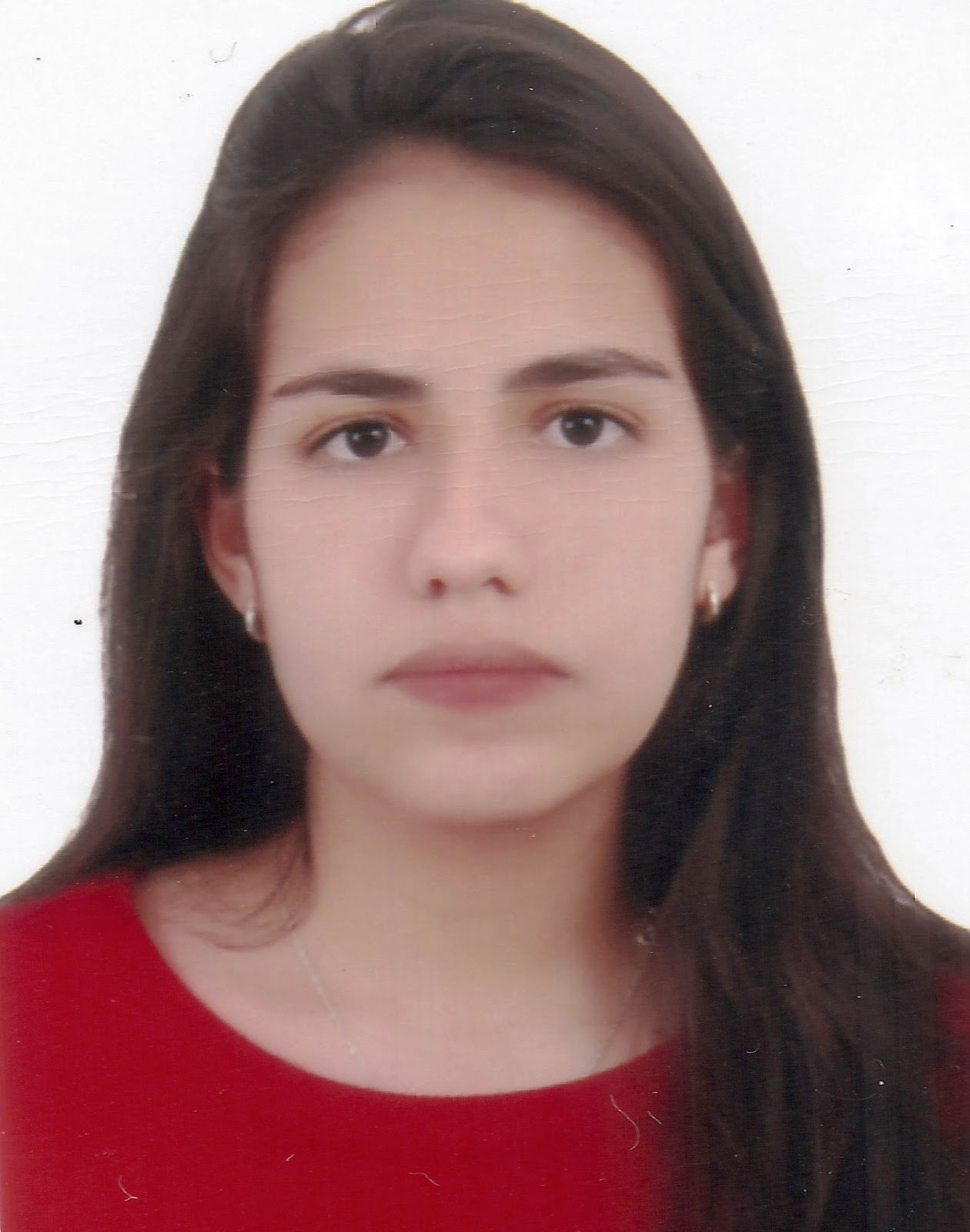}}]{Suzana Vilas Boas de Oliveira} Suzana Vilas Boas is currently pursuing a Ph.D. degree in the Signal Processing and Instrumentation Program at the University of São Paulo (EESC - USP) with a research focus on the development of a non-invasive motor imagery-based Brain-Computer Interface (MI-BCI) for the interaction with 3D images and automatic wheelchairs. She received her B.Sc. degree in Electrical Engineering, emphasis on Electronics and special studies in Biomedical Engineering, from the same institution. Her research interests include neuroscience, neuroplasticity, brain-computer interfaces, and artificial intelligence.
\end{IEEEbiography}

\begin{IEEEbiography}[{\includegraphics[width=1in,height=1.25in,clip]{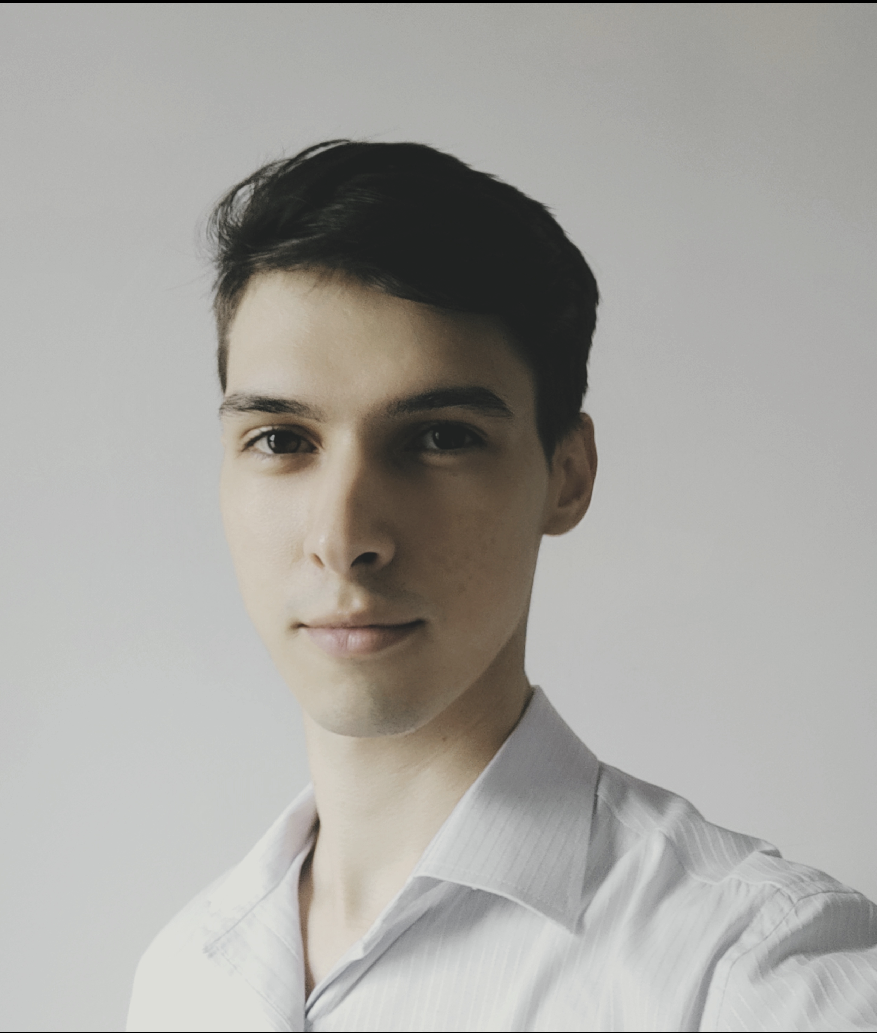}}]{Thiago Henrique Segreto Silva,}
Thiago H. Segreto received the B.S. degree in Mechatronics Engineering from the University of São Paulo, São Carlos, Brazil, in 2021, and the M.Sc. degree in Robotics with a specialization in Computer Vision from the same institution in 2025. He is currently pursuing a Ph.D. degree in Robotics, focusing on perception systems integrated with reinforcement learning. His research interests include robotic perception, deep reinforcement learning, computer vision, and autonomous systems, aiming at the development of intelligent, adaptive robots capable of operating effectively in complex, dynamic environments.
\end{IEEEbiography}

\begin{IEEEbiography}[{\includegraphics[width=1in,height=1.25in,clip]{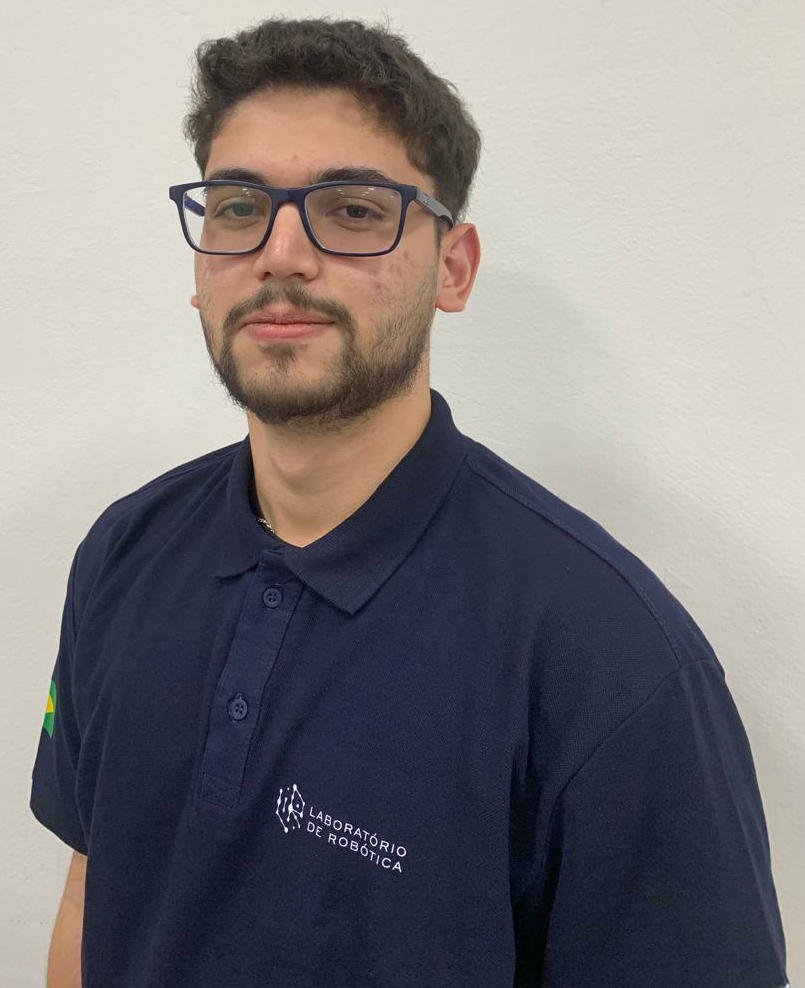}}]{Pedro Saraiva}
is currently pursuing a Bachelor of Engineering degree in Mechatronics Engineering at the University of São Paulo (USP). He is an undergraduate researcher within USP's Mobile Robotics Group, where he contributes to advancements in the field of mobile robotics, with a focus on applications for oil platforms.
\end{IEEEbiography}

\begin{IEEEbiography}[{\includegraphics[width=1in,height=1.25in,clip]{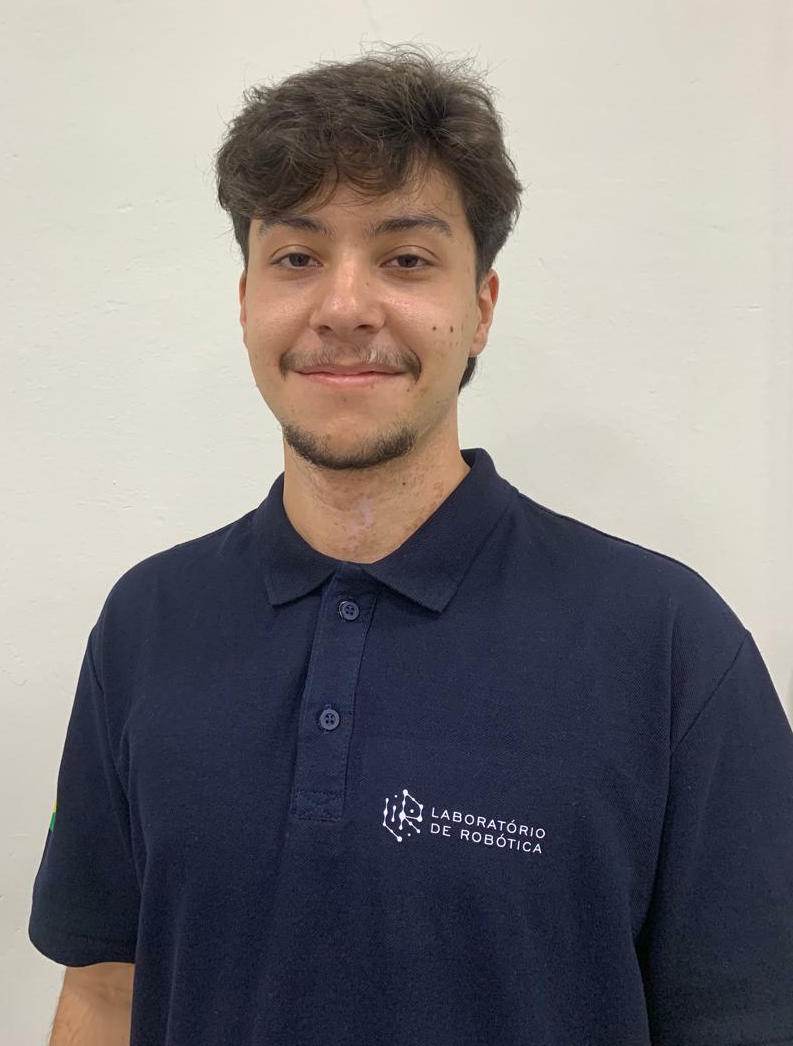}}]{Enzo Ferreira de Souza}
is currently pursuing a Bachelor of Engineering degree in Mechatronics Engineering at the University of São Paulo (USP). He is an undergraduate researcher within USP's Mobile Robotics Group, where he contributes to advancements in the field of mobile robotics, with a focus on applications for oil platforms.
\end{IEEEbiography}

\begin{IEEEbiography}[{\includegraphics[width=1in,height=1.25in,clip,keepaspectratio]{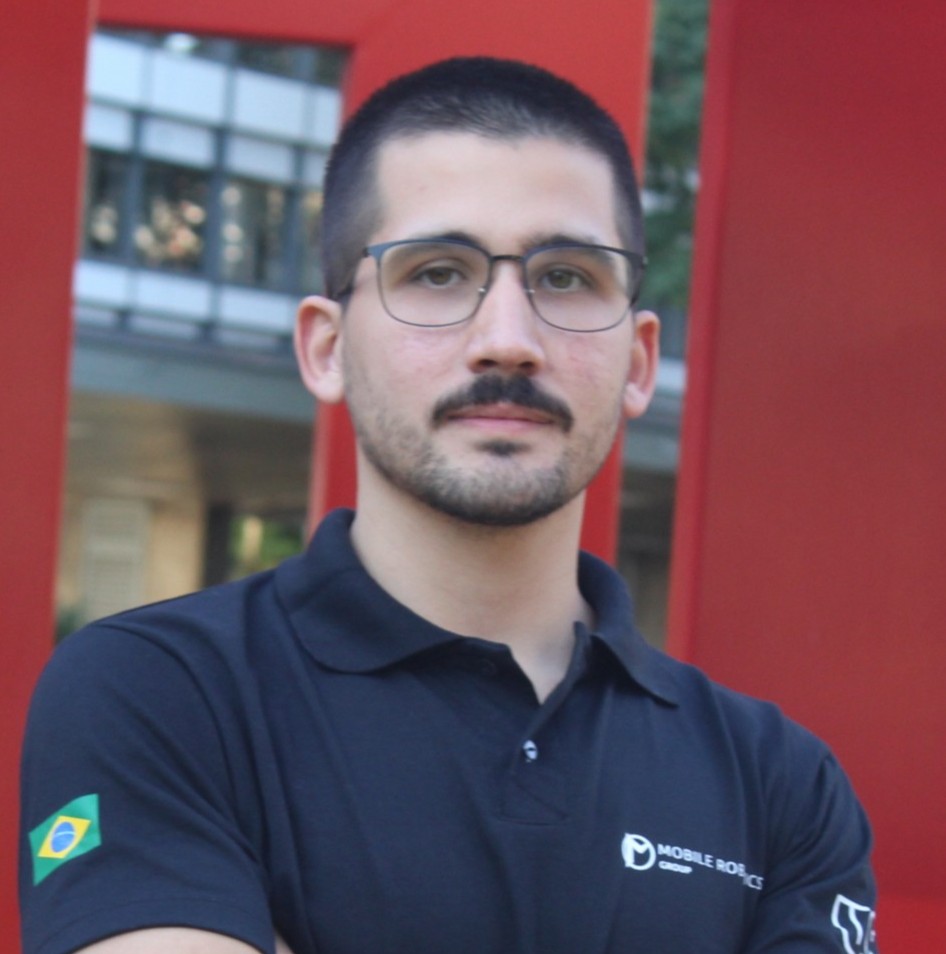}}]{Ricardo V. Godoy} received the Bachelor of Engineering in Mechatronics Engineering in 2019, followed by M.Sc. in 2021 in Mechanical Engineering from the University of São Paulo, São Carlos, Brazil. He received his PhD in Mechatronics Engineering with the New Dexterity Research Group of the University of Auckland, New Zealand, where he worked on the analysis and development of novel Human-Machine Interfaces (HMI) for the control of robotic and bionic devices while focusing on the challenges and limitations in the use of HMI for robust grasping and decoding of dexterous, in-hand manipulation tasks. He is currently pursuing his postdoc at the University of São Paulo, São Carlos, Brazil, working towards the development of robotic frameworks for inspection and automation, focusing on manipulation and loco-manipulation frameworks.
\end{IEEEbiography}

\begin{IEEEbiography}[{\includegraphics[width=1in,height=1.25in,clip,keepaspectratio]{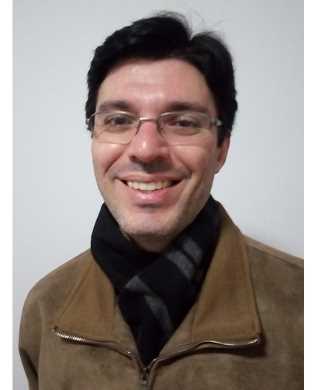}}]{Leonardo André Ambrosio} Leonardo A. Ambrosio received the B.Sc., M.Sc. and the Ph.D. degrees in Electrical Engineering from University of Campinas, Brazil, in 2002, 2005 and 2009, respectively. Between 2009 and 2013 he was a postdoctoral Fellow with the Department of Microwaves and Optics at the School of Electrical and Computer Engineering, University of Campinas, and developed part of his research at the University of Pennsylvania, Philadelphia, USA. He is currently an Associate Professor at University of São Paulo (SEL-EESC) and coordinates the Applied Electromagnetics Group (AEG). His research interests include photonics, light-scattering problems for optical trapping and manipulation, and modelling of non-diffracting beams envisioning applications in biomedical optics, telecommunications, holography, volumetric displays and atom guiding. He is also interested in brain-computer interfaces for entertainment, games and the metaverse, envisioning mind control of three-dimensional volumetric displays.
\end{IEEEbiography}

\begin{IEEEbiography}[{\includegraphics[width=1in,height=1.25in,clip,keepaspectratio]{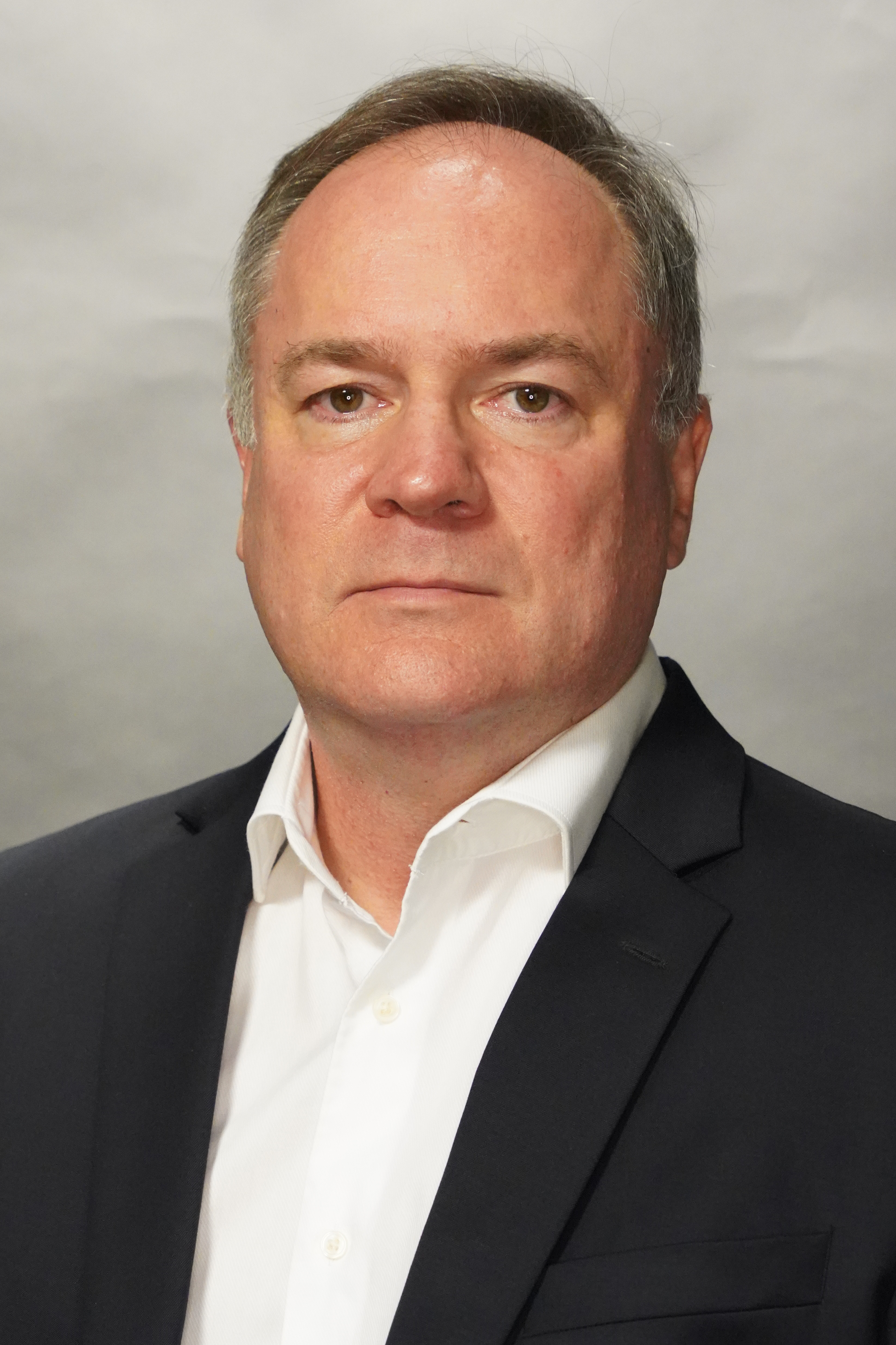}}]{Marcelo Becker} Marcelo Becker received the B.Sc. degree in Mechanical Engineering (Mechatronics) from the University of São Paulo, Brazil, in 1993, and the M.Sc. and D.Sc. degrees in Mechanical Engineering from the University of Campinas, Brazil, in 1997 and 2000, respectively. He was a visiting researcher at ETH Zürich and did a sabbatical at EPF Lausanne, Switzerland. He is currently an Associate Professor at the University of São Paulo and coordinates the Mobile Robotics Group and the USP Center of Robotics (CRob). His research interests include mobile robotics, automation, perception systems, and mechatronic design for applications in agriculture and industrial automation.
\end{IEEEbiography}

\onecolumn
\appendices
\section{Tables \& Figures Results}\label{appendix:A}
\input{apendice}

\EOD
\end{document}

%% file: apendice.tex
\begin{figure*}[htbp]
    \centering
    \includegraphics[width=1.0\textwidth]{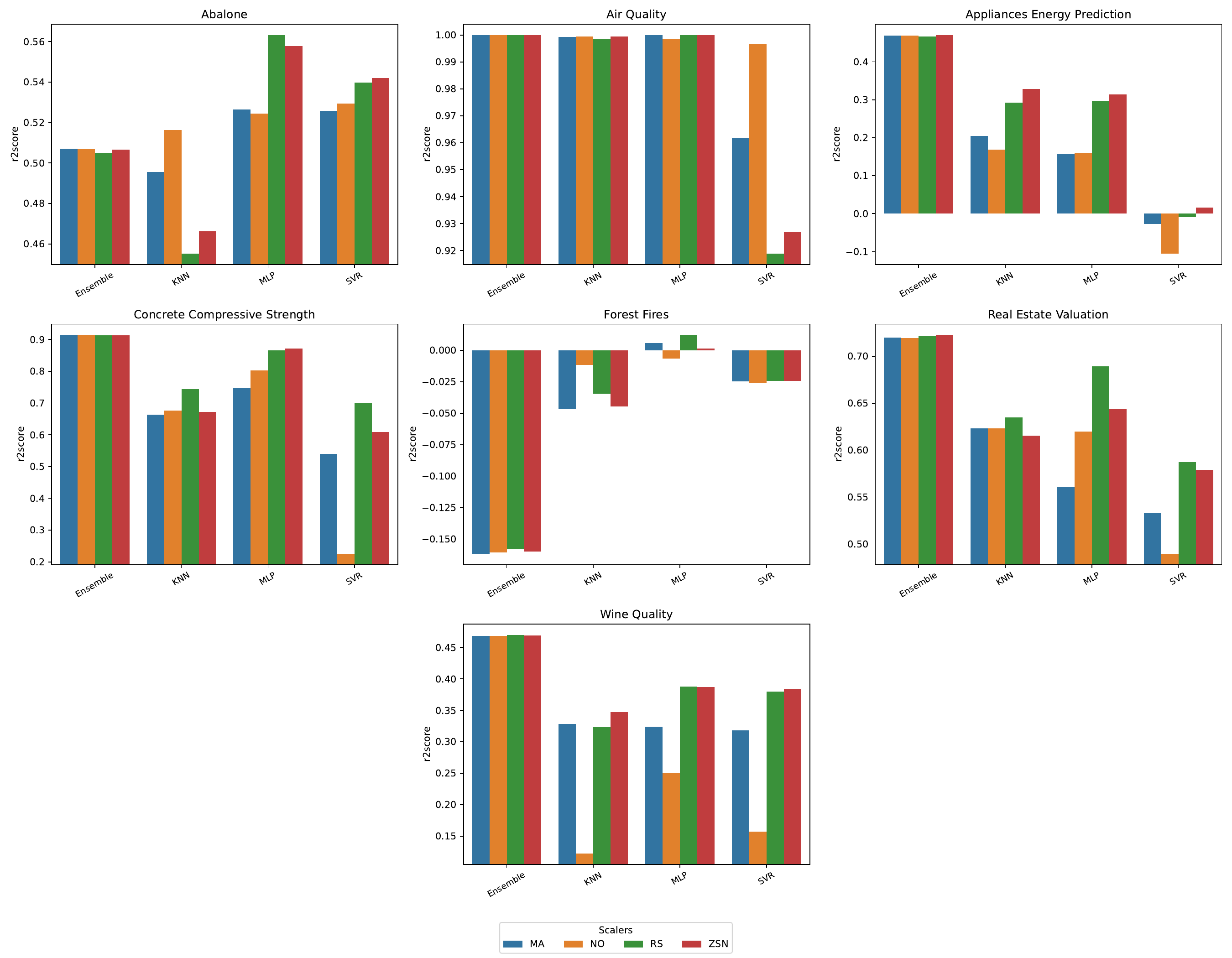}
    \caption{$R^2$ results for the 7 datasets.  
    The ensemble models are: RF, LGBM, CatBoost, and XGBoost.}
    \label{fig:all_datasets_r2}
\end{figure*}


\onecolumn
\begin{figure}[H]
    \centering
    \includegraphics[width=1.0\textwidth]{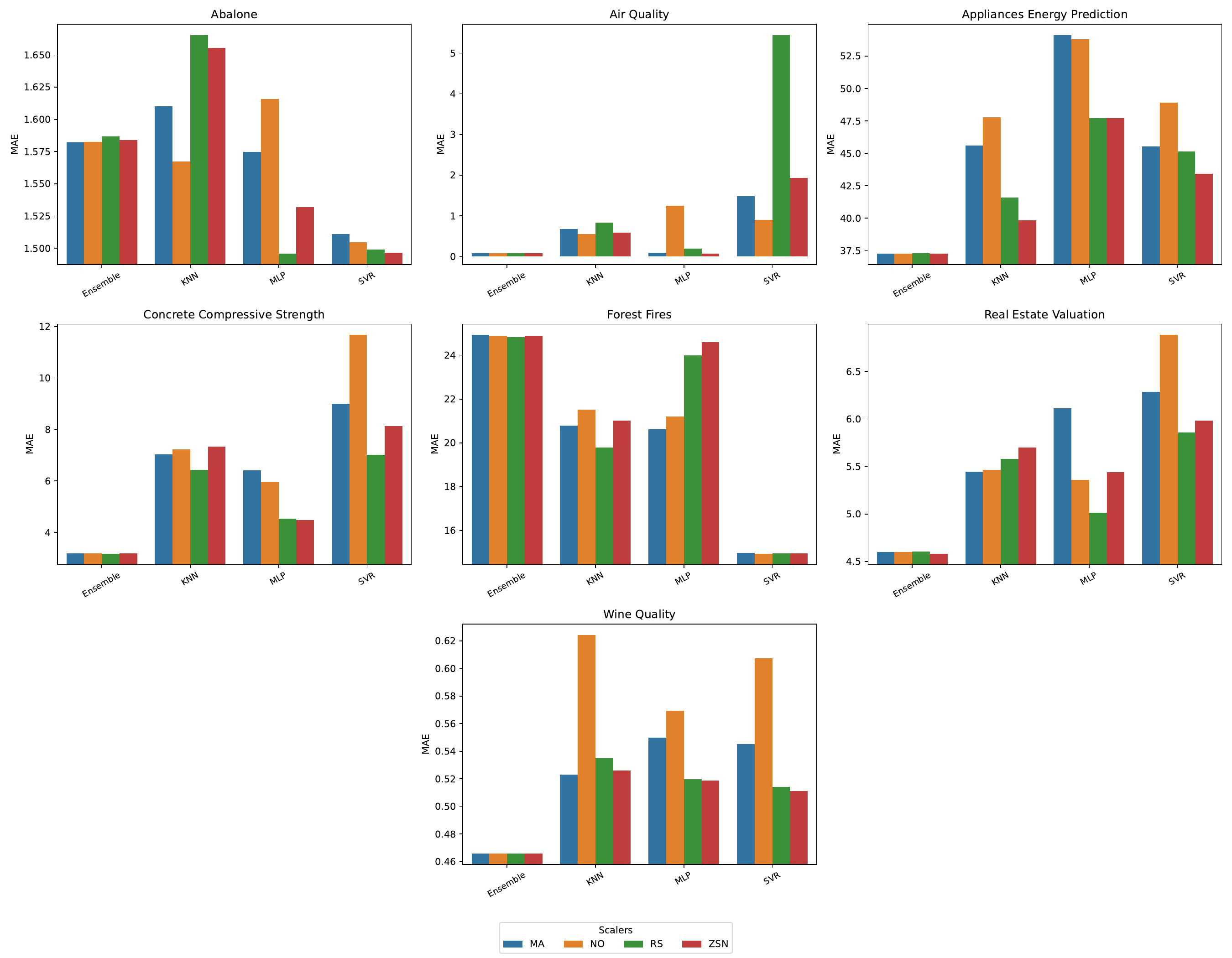}
    \caption{MAE results for the 7 datasets.  
    The ensemble models are: RF, LGBM, CatBoost, and XGBoost.}
    \label{fig:all_datasets_mae}
\end{figure}
\twocolumn
\onecolumn
\begin{figure}[H]
    \centering
    \includegraphics[width=1.0\textwidth]{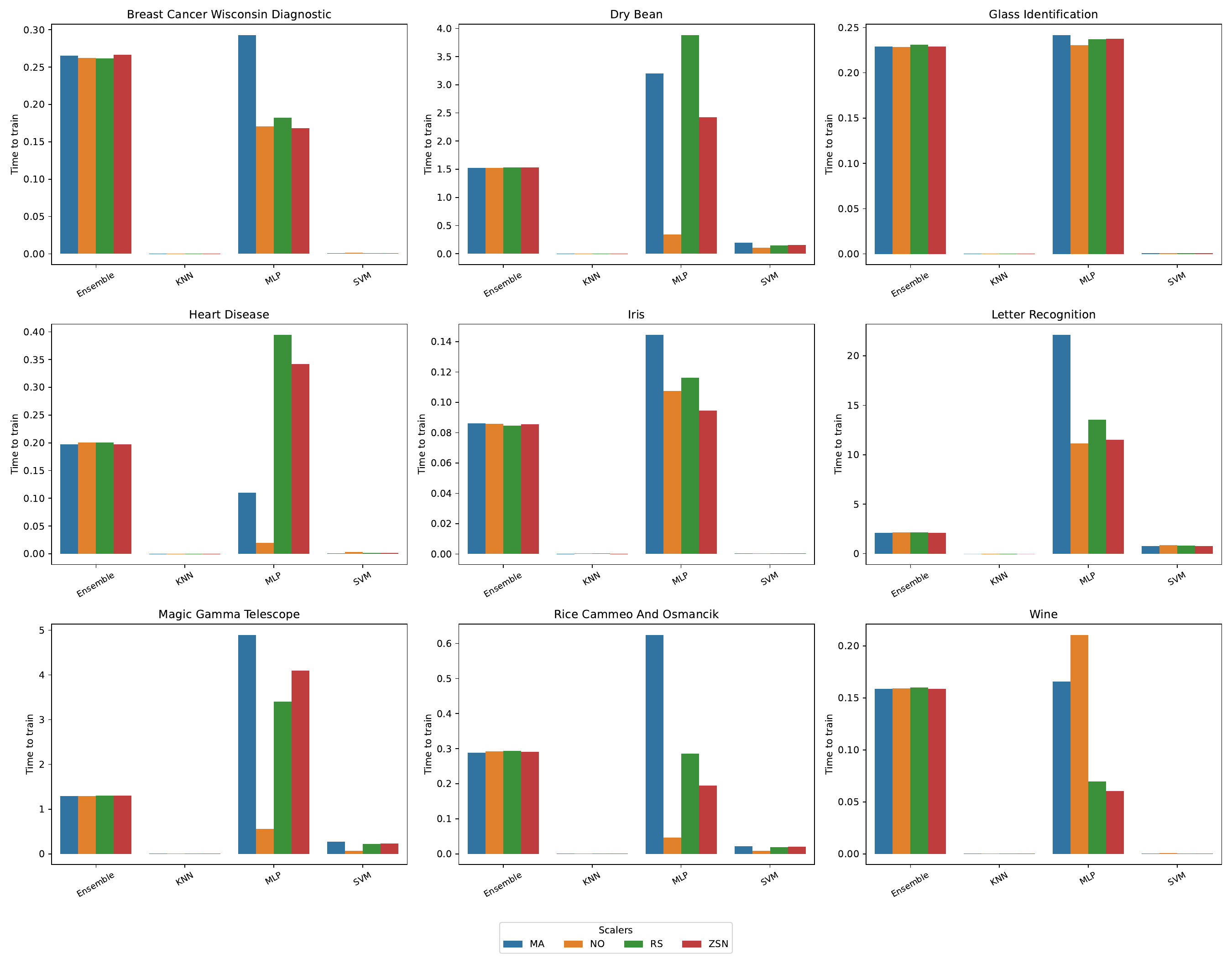}
    \caption{Time to train for the classification tasks. The ensemble models are: RF, LGBM, CatBoost, and XGBoost.}
    \label{fig:all_datasets_class_train}
\end{figure}
\twocolumn

\onecolumn
\begin{figure}[H]
    \centering
    \includegraphics[width=1.0\textwidth]{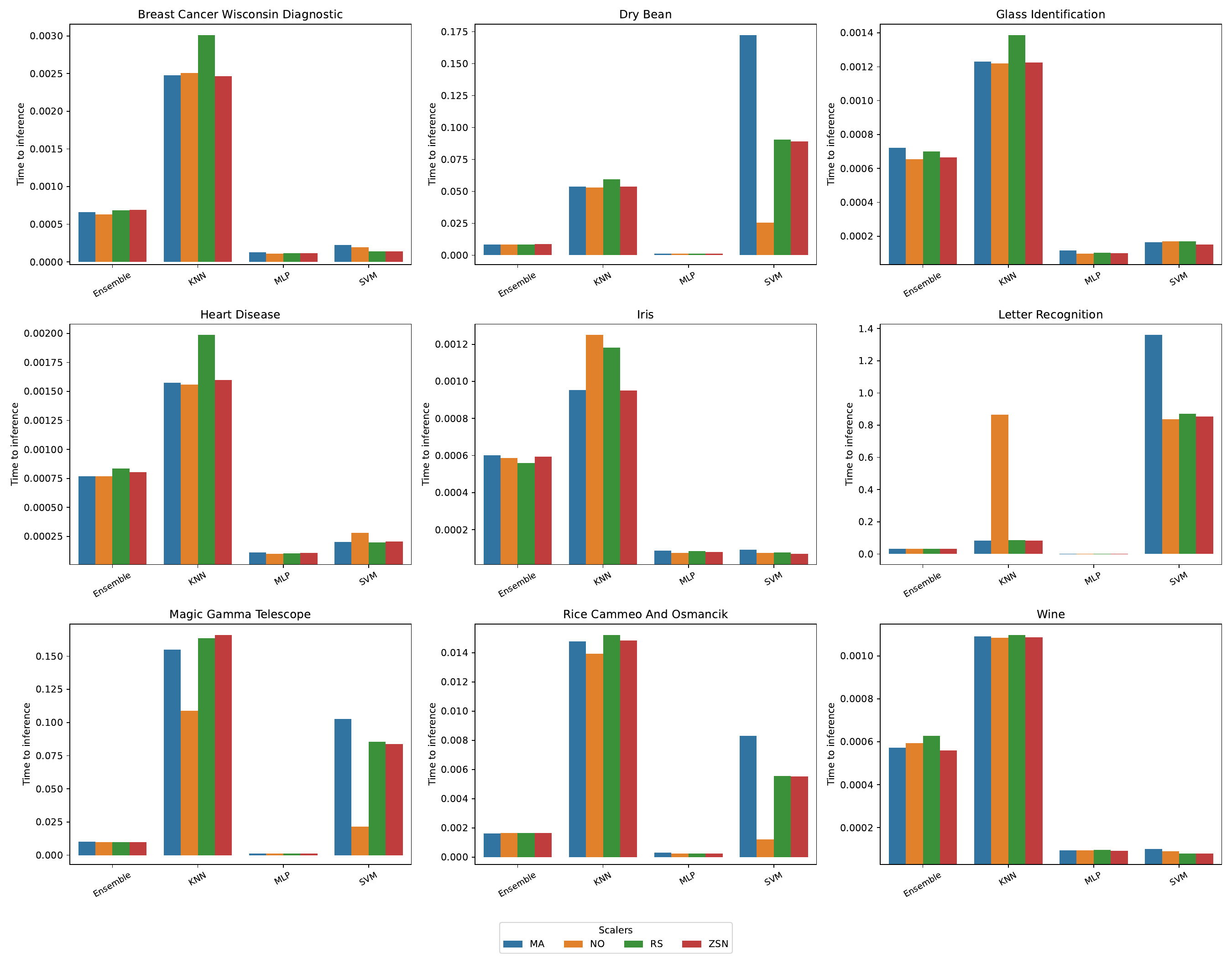}
    \caption{Time to inference for the classification tasks. The ensemble models are: RF, LGBM, CatBoost, and XGBoost.}
    \label{fig:all_datasets_class_inf}
\end{figure}
\twocolumn

\onecolumn
\begin{figure}[H]
    \centering
    \includegraphics[width=1.0\textwidth]{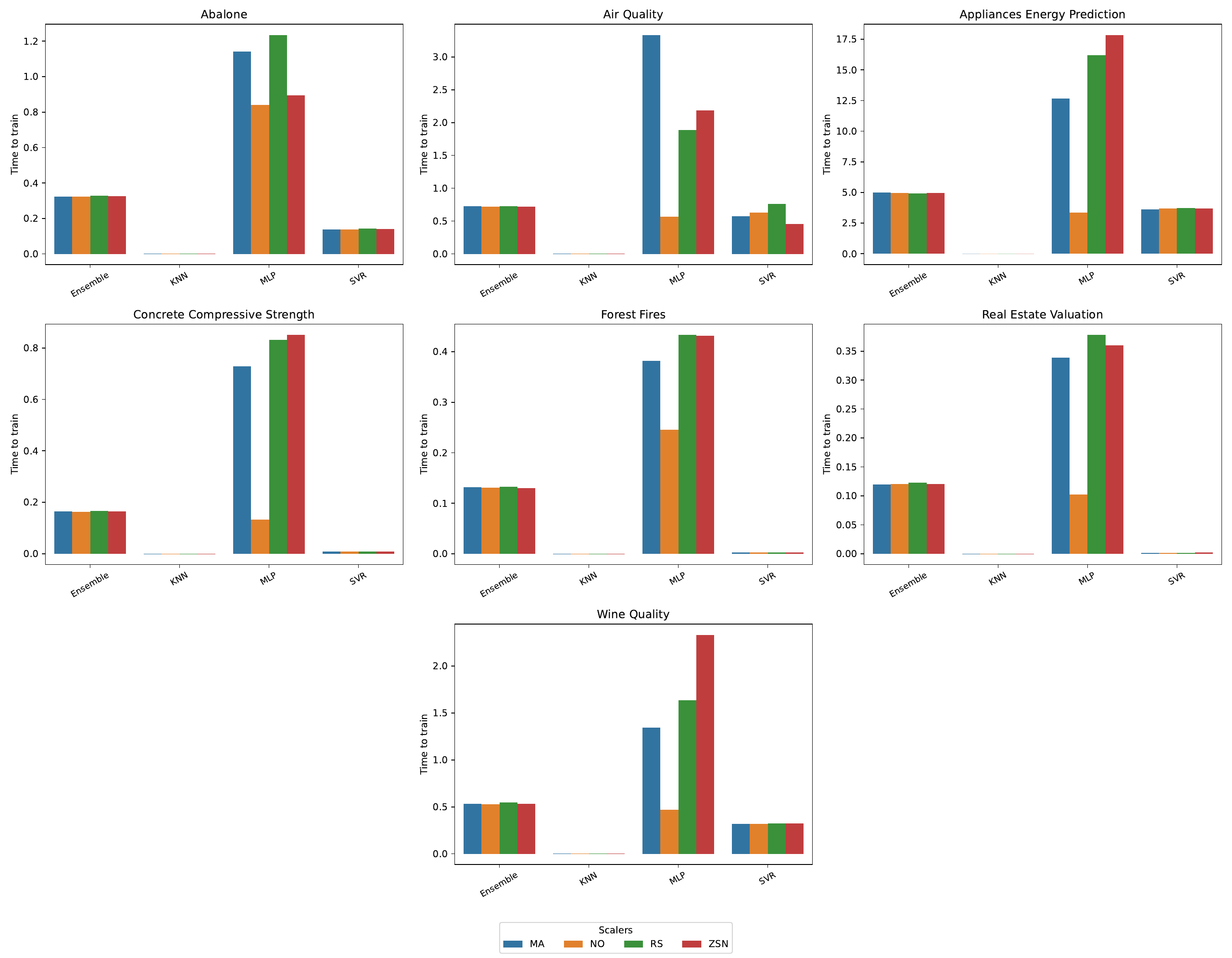}
    \caption{Time to train for the regression tasks. The ensemble models are: RF, LGBM, CatBoost, and XGBoost.}
    \label{fig:all_datasets_ref_train}
\end{figure}
\twocolumn
\onecolumn
\begin{figure}[H]
    \centering
    \includegraphics[width=1.0\textwidth]{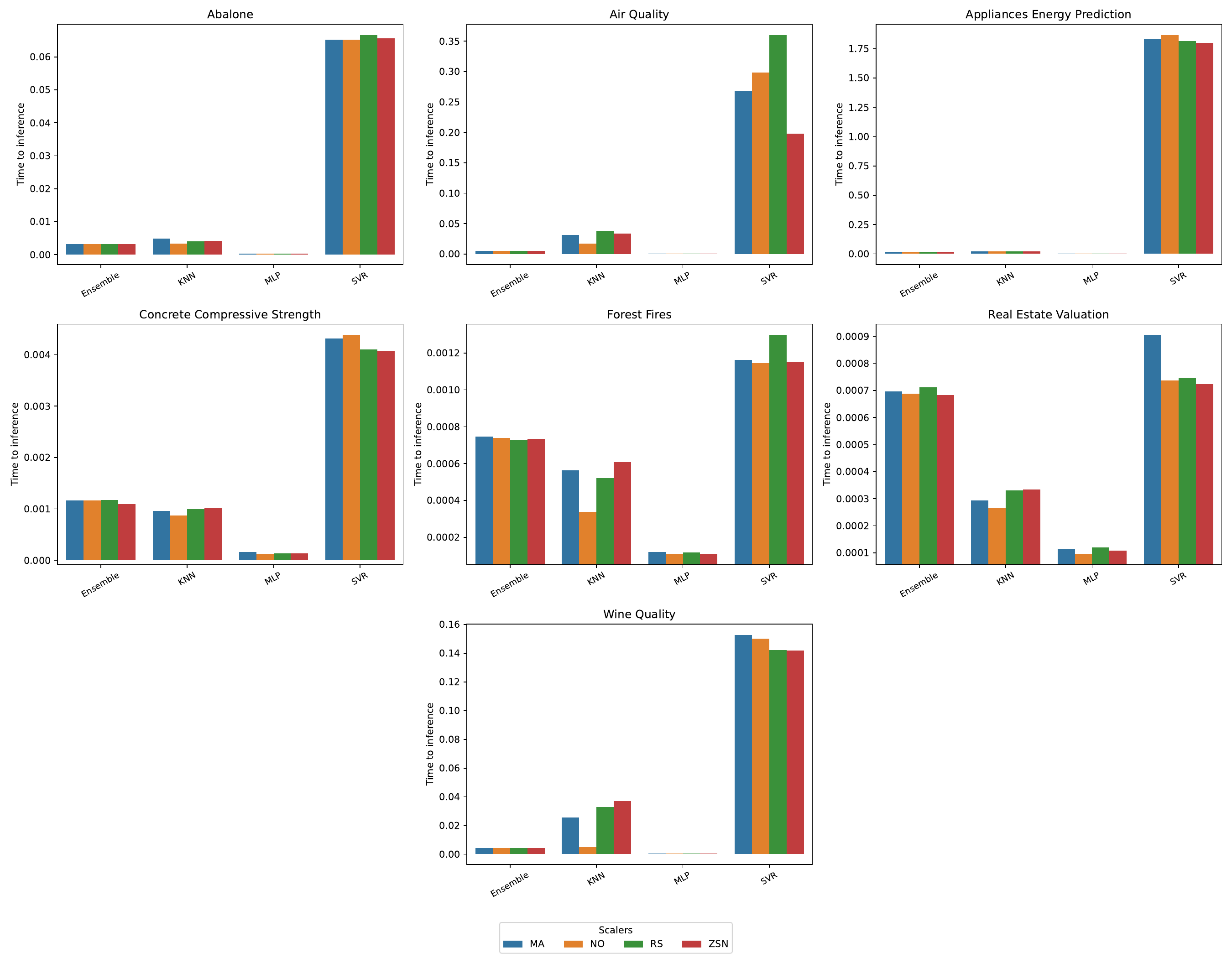}
    \caption{Time to inference for the regression tasks. The ensemble models are: RF, LGBM, CatBoost, and XGBoost.}
    \label{fig:all_datasets_reg_inf}
\end{figure}
\twocolumn
\onecolumn
\begin{figure}[H]
    \centering
    \includegraphics[width=0.95\textwidth]{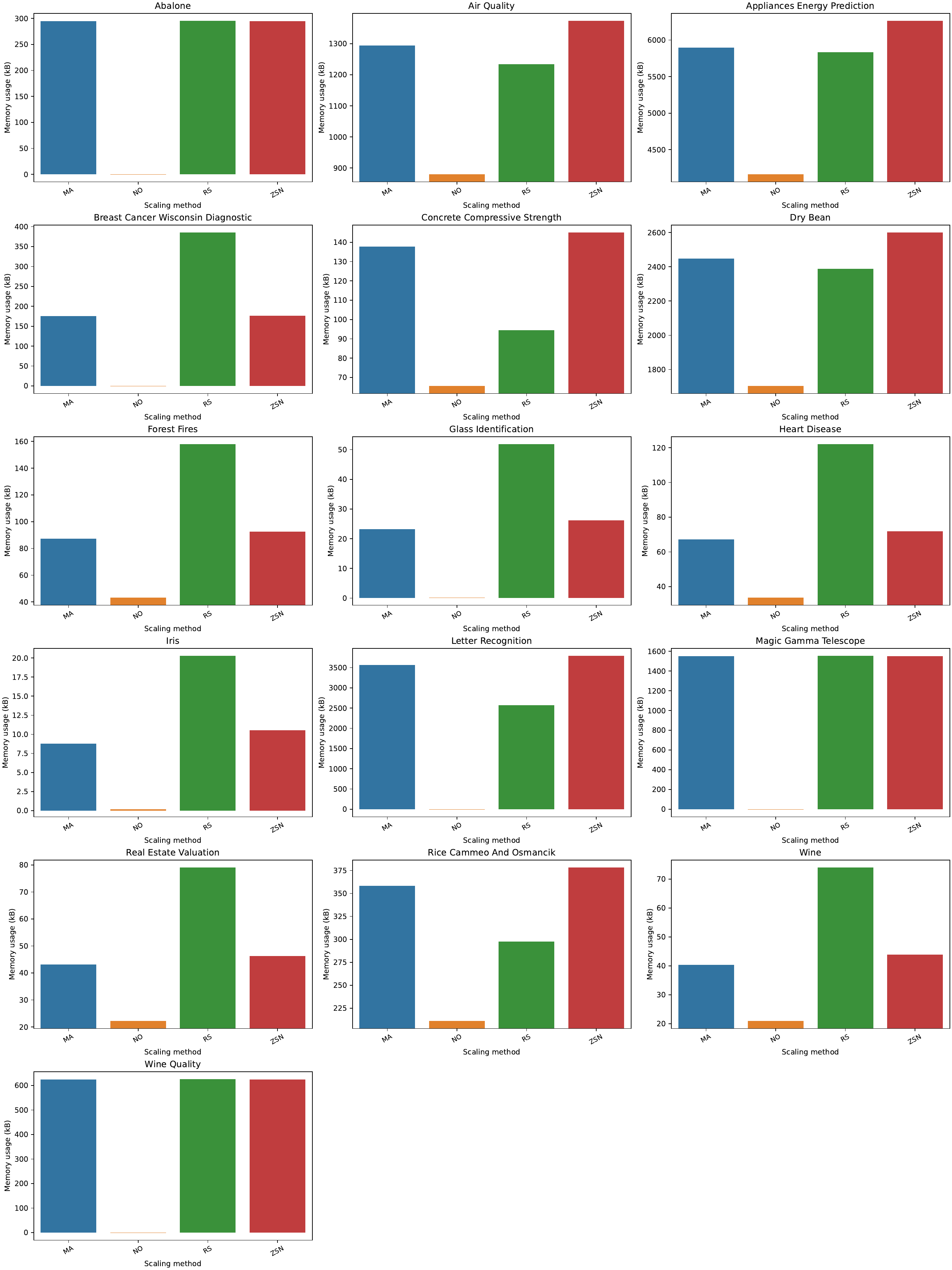}
    \caption{Memory Usage (kB) per Dataset and Scaling Method}
    \label{fig:all_datasets_mem}
\end{figure}

{\footnotesize
\setlength{\tabcolsep}{2pt} 
\begin{longtable}{lrrrrrr}
\caption{Wilcoxon test results (scaled vs. unscaled features) and Friedman test results (across all scaling methods) for accuracy.} \label{tab:accuracy_stests} \\
\toprule
Model & Wilcoxon stat & Wilcoxon $p$ & Wilcoxon sig. & Friedman stat & Friedman $p$ & Friedman sig. \\
\midrule
\endfirsthead
\caption[]{Wilcoxon test results (scaled vs. unscaled features) and Friedman test results (across all scaling methods) for accuracy.} \\
\toprule
Model & Wilcoxon stat & Wilcoxon $p$ & Wilcoxon sig. & Friedman stat & Friedman $p$ & Friedman sig. \\
\midrule
\endhead
\midrule
\multicolumn{7}{r}{Continued on next page} \\
\midrule
\endfoot
\bottomrule
\endlastfoot
LR & 1927.5000 & 0.3228 & No & 44.9579 & 0.0000 & Yes \\
SVM & 1392.0000 & 0.0001 & Yes & 42.0466 & 0.0000 & Yes \\
MLP & 744.5000 & 0.0000 & Yes & 34.4050 & 0.0006 & Yes \\
RF & 368.0000 & 0.9710 & No & 10.2824 & 0.5912 & No \\
NB & 706.0000 & 0.3379 & No & 5.7230 & 0.9294 & No \\
CART & 164.0000 & 0.0372 & No & 5.5087 & 0.9388 & No \\
LGBM & 336.0000 & 0.0252 & No & 17.3499 & 0.1369 & No \\
Ada & 0.0000 & 0.3173 & No & 12.0000 & 0.4457 & No \\
CatBoost & 31.0000 & 0.0913 & No & 9.7680 & 0.6363 & No \\
XGBoost & 19.0000 & 0.0329 & No & 5.9406 & 0.9190 & No \\
KNN & 372.0000 & 0.0000 & Yes & 28.9849 & 0.0040 & Yes \\
TabNet & 2312.5000 & 0.0732 & No & 17.1439 & 0.1443 & No \\
\end{longtable}
}

{\footnotesize
\setlength{\tabcolsep}{2pt} 
\begin{longtable}{lrrrrrrrr}
\caption{Wilcoxon test results (scaled vs. unscaled features) and Friedman test results (across all scaling methods) for $R^2$.} \label{tab:r2_stests} \\
\toprule
Model & Wilcoxon stat & Wilcoxon $p$ & Wilcoxon sig. & Friedman stat & Friedman $p$ & Friedman sig. \\
\midrule
\endfirsthead
\caption[]{Wilcoxon test results (scaled vs. unscaled features) and Friedman test results (across all scaling methods) for $R^2$.} \\
\toprule
Model & Wilcoxon stat & Wilcoxon $p$ & Wilcoxon sig. & Friedman stat & Friedman $p$ & Friedman sig. \\
\midrule
\endhead
\midrule
\multicolumn{7}{r}{Continued on next page} \\
\midrule
\endfoot
\bottomrule
\endlastfoot
SVR & 574.0000 & 0.0000 & Yes & 33.9245 & 0.0007 & Yes \\
MLP & 1193.0000 & 0.0083 & Yes & 47.3783 & 0.0000 & Yes \\
RF & 1721.0000 & 0.7753 & No & 13.2075 & 0.3541 & No \\
CART & 1152.0000 & 0.0530 & No & 5.9805 & 0.9171 & No \\
LGBM & 740.0000 & 0.0127 & No & 0.4284 & 1.0000 & No \\
Ada & 732.0000 & 0.0000 & Yes & 18.9670 & 0.0893 & No \\
CatBoost & 1463.0000 & 0.1510 & No & 2.5755 & 0.9979 & No \\
LinearRegression & 896.0000 & 0.1806 & No & 13.7553 & 0.3166 & No \\
XGBoost & 4.0000 & 0.7150 & No & 18.0000 & 0.1157 & No \\
KNN & 1275.0000 & 0.1860 & No & 21.8466 & 0.0393 & No \\
TabNet & 537.0000 & 0.0000 & Yes & 17.1805 & 0.1429 & No \\
\end{longtable}
}

{\footnotesize
\setlength{\tabcolsep}{2pt} 
\begin{longtable}{lrrrrrrrr}
\caption{Wilcoxon test results (scaled vs. unscaled features) and Friedman test results (across all scaling methods) for MSE.} \label{tab:mse_stests} \\
\toprule
Model & Wilcoxon stat & Wilcoxon $p$ & Wilcoxon sig. & Friedman stat & Friedman $p$ & Friedman sig. \\
\midrule
\endfirsthead
\caption[]{Wilcoxon test results (scaled vs. unscaled features) and Friedman test results (across all scaling methods) for MSE.} \\
\toprule
Model & Wilcoxon stat & Wilcoxon $p$ & Wilcoxon sig. & Friedman stat & Friedman $p$ & Friedman sig. \\
\midrule
\endhead
\midrule
\multicolumn{7}{r}{Continued on next page} \\
\midrule
\endfoot
\bottomrule
\endlastfoot
SVR & 826.0000 & 0.0000 & Yes & 33.9245 & 0.0007 & Yes \\
MLP & 1187.0000 & 0.0077 & Yes & 47.3783 & 0.0000 & Yes \\
RF & 1650.0000 & 0.5471 & No & 13.2075 & 0.3541 & No \\
CART & 1209.0000 & 0.0987 & No & 5.9805 & 0.9171 & No \\
LGBM & 875.0000 & 0.0991 & No & 0.4284 & 1.0000 & No \\
Ada & 794.0000 & 0.0000 & Yes & 18.9670 & 0.0893 & No \\
CatBoost & 1347.0000 & 0.0508 & No & 2.5755 & 0.9979 & No \\
LinearRegression & 1273.5000 & 0.4236 & No & 13.7403 & 0.3176 & No \\
XGBoost & 4.0000 & 0.7150 & No & 18.0000 & 0.1157 & No \\
KNN & 1458.0000 & 0.6811 & No & 21.8466 & 0.0393 & No \\
TabNet & 811.0000 & 0.0000 & Yes & 17.1805 & 0.1429 & No \\
\end{longtable}
}
{\footnotesize
\setlength{\tabcolsep}{2pt} 
\begin{longtable}{lrrrrrrrr}
\caption{Wilcoxon test results (scaled vs. unscaled features) and Friedman test results (across all scaling methods) for MAE.} \label{tab:mae_stests} \\
\toprule
Model & Wilcoxon stat & Wilcoxon $p$ & Wilcoxon sig. & Friedman stat & Friedman $p$ & Friedman sig. \\
\midrule
\endfirsthead
\caption[]{Wilcoxon test results (scaled vs. unscaled features) and Friedman test results (across all scaling methods) for MAE.} \\
\toprule
Model & Wilcoxon stat & Wilcoxon $p$ & Wilcoxon sig. & Friedman stat & Friedman $p$ & Friedman sig. \\
\midrule
\endhead
\midrule
\multicolumn{7}{r}{Continued on next page} \\
\midrule
\endfoot
\bottomrule
\endlastfoot
SVR & 885.0000 & 0.0001 & Yes & 19.3019 & 0.0815 & No \\
MLP & 1382.0000 & 0.0723 & No & 32.5526 & 0.0011 & Yes \\
RF & 1234.0000 & 0.0140 & No & 6.9434 & 0.8613 & No \\
CART & 1217.5000 & 0.2036 & No & 5.5396 & 0.9375 & No \\
LGBM & 1138.0000 & 0.9950 & No & 5.3525 & 0.9452 & No \\
Ada & 1171.0000 & 0.0212 & No & 20.7147 & 0.0547 & No \\
CatBoost & 1255.0000 & 0.0181 & No & 9.9273 & 0.6223 & No \\
LinearRegression & 1237.5000 & 0.6677 & No & 6.0974 & 0.9111 & No \\
XGBoost & 4.0000 & 0.7150 & No & 18.0000 & 0.1157 & No \\
KNN & 1091.0000 & 0.0252 & No & 11.2601 & 0.5068 & No \\
TabNet & 697.0000 & 0.0000 & Yes & 13.6954 & 0.3206 & No \\
\end{longtable}
}

{\footnotesize
\setlength{\tabcolsep}{2pt} 
\begin{longtable}{lrrrrrrrr}
\caption{Wilcoxon test results (scaled vs. unscaled features) and Friedman test results (across all scaling methods) for time to train.} \label{tab:training_stests} \\
\toprule
Model & Wilcoxon stat & Wilcoxon $p$ & Wilcoxon sig. & Friedman stat & Friedman $p$ & Friedman sig. \\
\midrule
\endfirsthead
\caption[]{Wilcoxon test results (scaled vs. unscaled features) and Friedman test results (across all scaling methods) for time to train.} \\
\toprule
Model & Wilcoxon stat & Wilcoxon $p$ & Wilcoxon sig. & Friedman stat & Friedman $p$ & Friedman sig. \\
\midrule
\endhead
\midrule
\multicolumn{7}{r}{Continued on next page} \\
\midrule
\endfoot
\bottomrule
\endlastfoot
LR & 648.5000 & 0.0000 & Yes & 39.2741 & 0.0001 & Yes \\
SVM & 1424.5000 & 0.0000 & Yes & 49.5541 & 0.0000 & Yes \\
MLP & 4831.5000 & 0.0000 & Yes & 44.8613 & 0.0000 & Yes \\
RF & 6612.5000 & 0.0006 & Yes & 14.2508 & 0.2850 & No \\
NB & 2654.0000 & 0.9371 & No & 21.3755 & 0.0451 & No \\
CART & 9137.0000 & 0.9677 & No & 21.1547 & 0.0482 & No \\
LGBM & 8306.5000 & 0.2143 & No & 15.7088 & 0.2049 & No \\
Ada & 6118.0000 & 0.0000 & Yes & 14.1374 & 0.2920 & No \\
CatBoost & 8183.0000 & 0.1609 & No & 8.5467 & 0.7411 & No \\
XGBoost & 4454.0000 & 0.0000 & Yes & 37.8845 & 0.0002 & Yes \\
KNN & 5243.5000 & 0.0000 & Yes & 84.2967 & 0.0000 & Yes \\
TabNet & 7290.0000 & 0.0105 & No & 21.1319 & 0.0485 & No \\
SVR & 494.0000 & 0.0000 & Yes & 45.6641 & 0.0000 & Yes \\
LinearRegression & 211.0000 & 0.0000 & Yes & 42.2801 & 0.0000 & Yes \\
\end{longtable}
}

{\footnotesize
\setlength{\tabcolsep}{2pt} 
\begin{longtable}{lrrrrrrrr}
\caption{Wilcoxon test results (scaled vs. unscaled features) and Friedman test results (across all scaling methods) for time to inference.} \label{tab:inference_stests} \\
\toprule
Model & Wilcoxon stat & Wilcoxon $p$ & Wilcoxon sig. & Friedman stat & Friedman $p$ & Friedman sig. \\
\midrule
\endfirsthead
\caption[]{Wilcoxon test results (scaled vs. unscaled features) and Friedman test results (across all scaling methods) for time to inference.} \\
\toprule
Model & Wilcoxon stat & Wilcoxon $p$ & Wilcoxon sig. & Friedman stat & Friedman $p$ & Friedman sig. \\
\midrule
\endhead
\midrule
\multicolumn{7}{r}{Continued on next page} \\
\midrule
\endfoot
\bottomrule
\endlastfoot
LR & 648.5000 & 0.0000 & Yes & 39.2741 & 0.0001 & Yes \\
SVM & 1424.5000 & 0.0000 & Yes & 49.5541 & 0.0000 & Yes \\
MLP & 4831.5000 & 0.0000 & Yes & 44.8613 & 0.0000 & Yes \\
RF & 6612.5000 & 0.0006 & Yes & 14.2508 & 0.2850 & No \\
NB & 2654.0000 & 0.9371 & No & 21.3755 & 0.0451 & No \\
CART & 9137.0000 & 0.9677 & No & 21.1547 & 0.0482 & No \\
LGBM & 8306.5000 & 0.2143 & No & 15.7088 & 0.2049 & No \\
Ada & 6118.0000 & 0.0000 & Yes & 14.1374 & 0.2920 & No \\
CatBoost & 8183.0000 & 0.1609 & No & 8.5467 & 0.7411 & No \\
XGBoost & 4454.0000 & 0.0000 & Yes & 37.8845 & 0.0002 & Yes \\
KNN & 5243.5000 & 0.0000 & Yes & 84.2967 & 0.0000 & Yes \\
TabNet & 7290.0000 & 0.0105 & No & 21.1319 & 0.0485 & No \\
SVR & 494.0000 & 0.0000 & Yes & 45.6641 & 0.0000 & Yes \\
LinearRegression & 211.0000 & 0.0000 & Yes & 42.2801 & 0.0000 & Yes \\
\end{longtable}
}

{\footnotesize
\setlength{\tabcolsep}{3pt} 
\begin{longtable}{llrrrrrrrrrrrrr}
\caption{Accuracy by dataset, model, and scaling method.} \label{tab:accuracy_big_table} \\
\toprule
 Dataset & Model & NO & MM & MA & ZSN & PS & VAST & MC & RS & QT & DS & TT & LS & HT \\
\midrule
\endfirsthead
\caption[]{Accuracy by dataset, model, and scaling method.} \\
\toprule
Dataset & Model & NO & MM & MA & ZSN & PS & VAST & MC & RS & QT & DS & TT & LS & HT \\
\midrule
\endhead
\midrule
\multicolumn{15}{r}{Continued on next page} \\
\midrule
\endfoot
\bottomrule
\endlastfoot
\multirow[t]{12}{*}{\begin{tabular}[t]{@{}l@{}}Breast Cancer \\ Wisconsin \\ Diagnostic\end{tabular}} & Ada & 0.9766 & 0.9766 & 0.9766 & 0.9766 & 0.9766 & 0.9766 & 0.9766 & 0.9766 & 0.9766 & 0.9766 & 0.9766 & 0.9766 & 0.9766 \\
 & CART & 0.9415 & 0.9415 & 0.9415 & 0.9415 & 0.9415 & 0.9415 & 0.9415 & 0.9415 & 0.9474 & 0.9415 & 0.9415 & 0.9474 & 0.9474 \\
 & CatBoost & 0.9766 & 0.9766 & 0.9766 & 0.9766 & 0.9766 & 0.9766 & 0.9766 & 0.9766 & 0.9766 & 0.9766 & 0.9766 & 0.9766 & 0.9766 \\
 & KNN & 0.9591 & 0.9649 & 0.9766 & 0.9591 & 0.9766 & 0.9591 & 0.9591 & 0.9649 & 0.9649 & 0.9474 & 0.9591 & 0.9591 & 0.9591 \\
 & LGBM & 0.9474 & 0.9474 & 0.9474 & 0.9591 & 0.9591 & 0.9591 & 0.9591 & 0.9591 & 0.9474 & 0.9474 & 0.9415 & 0.9415 & 0.9591 \\
 & LR & 0.9708 & 0.9649 & 0.9532 & 0.9825 & 0.9766 & 0.9883 & 0.9708 & 0.9883 & 0.9942 & 0.9123 & 0.6316 & 0.9883 & 0.9942 \\
 & MLP & 0.9649 & 0.9766 & 0.9766 & 0.9766 & 0.9883 & 0.9708 & 0.9006 & 0.9708 & 0.9766 & 0.9825 & 0.6316 & 0.9825 & 0.9883 \\
 & NB & 0.9415 & 0.9357 & 0.9357 & 0.9357 & 0.9357 & 0.9357 & 0.9415 & 0.9357 & 0.9532 & 0.9357 & 0.9357 & 0.9532 & 0.9532 \\
\end{longtable}
}
{\footnotesize
\setlength{\tabcolsep}{2pt} 
\begin{longtable}{llrrrrrrrrrrrrr}
\caption{Time to Train (s) by dataset, model, and scaling method.} \label{tab:train_class_big_table} \\
\toprule
Dataset & Model & NO & MM & MA & ZSN & PS & VAST & MC & RS & QT & DS & TT & LS & HT \\
\midrule
\endfirsthead
\caption[]{Time to Train (s) by dataset, model, and scaling method.} \\
\toprule
Dataset & Model & NO & MM & MA & ZSN & PS & VAST & MC & RS & QT & DS & TT & LS & HT \\
\midrule
\endhead
\midrule
\multicolumn{15}{r}{Continued on next page} \\
\midrule
\endfoot
\bottomrule
\endlastfoot
\multirow[t]{12}{*}{\begin{tabular}[t]{@{}l@{}}Breast Cancer \\ Wisconsin \\ Diagnostic\end{tabular}} & Ada & 0.0625 & 0.0660 & 0.0623 & 0.0626 & 0.0661 & 0.0619 & 0.0618 & 0.0617 & 0.0616 & 0.0618 & 0.0621 & 0.0619 & 0.0618 \\
 & CART & 0.0035 & 0.0036 & 0.0039 & 0.0035 & 0.0040 & 0.0036 & 0.0035 & 0.0035 & 0.0035 & 0.0035 & 0.0036 & 0.0036 & 0.0036 \\
 & CatBoost & 0.9431 & 0.9772 & 0.9543 & 0.9512 & 0.9513 & 0.9340 & 0.9568 & 0.9348 & 0.9510 & 0.9467 & 0.9513 & 0.9505 & 0.9479 \\
 & KNN & 0.0002 & 0.0002 & 0.0002 & 0.0002 & 0.0002 & 0.0002 & 0.0002 & 0.0002 & 0.0002 & 0.0002 & 0.0002 & 0.0002 & 0.0003 \\
 & LGBM & 0.0213 & 0.0306 & 0.0219 & 0.0275 & 0.0237 & 0.0233 & 0.0228 & 0.0256 & 0.0213 & 0.0210 & 0.0216 & 0.0213 & 0.0239 \\
 & LR & 0.0043 & 0.0047 & 0.0014 & 0.0012 & 0.0025 & 0.0024 & 0.0044 & 0.0012 & 0.0012 & 0.0011 & 0.0013 & 0.0012 & 0.0010 \\
 & MLP & 0.1708 & 0.2342 & 0.2929 & 0.1683 & 0.2549 & 0.1412 & 0.0107 & 0.1824 & 0.3734 & 0.3249 & 0.0070 & 0.3637 & 0.1990 \\
 & NB & 0.0003 & 0.0003 & 0.0003 & 0.0003 & 0.0004 & 0.0003 & 0.0004 & 0.0003 & 0.0003 & 0.0004 & 0.0003 & 0.0004 & 0.0004 \\
\end{longtable}
}
{\footnotesize
\setlength{\tabcolsep}{3pt} 
\begin{longtable}{llrrrrrrrrrrrrr}
\caption{Time to Inference (s) by dataset, model, and scaling method.} \label{tab:inf_class_big_table} \\
\toprule
Dataset & Model & NO & MM & MA & ZSN & PS & VAST & MC & RS & QT & DS & TT & LS & HT \\
\midrule
\endfirsthead
\caption[]{Time to Inference (s) by dataset, model, and scaling method.} \\
\toprule
Dataset & Model & NO & MM & MA & ZSN & PS & VAST & MC & RS & QT & DS & TT & LS & HT \\
\midrule
\endhead
\midrule
\multicolumn{15}{r}{Continued on next page} \\
\midrule
\endfoot
\bottomrule
\endlastfoot
\multirow[t]{12}{*}{\begin{tabular}[t]{@{}l@{}}Breast Cancer \\ Wisconsin \\ Diagnostic\end{tabular}} & Ada & 0.0018 & 0.0019 & 0.0018 & 0.0018 & 0.0018 & 0.0018 & 0.0018 & 0.0018 & 0.0018 & 0.0018 & 0.0018 & 0.0018 & 0.0018 \\
 & CART & 0.0001 & 0.0001 & 0.0001 & 0.0001 & 0.0001 & 0.0001 & 0.0001 & 0.0001 & 0.0001 & 0.0001 & 0.0001 & 0.0001 & 0.0001 \\
 & CatBoost & 0.0005 & 0.0005 & 0.0006 & 0.0004 & 0.0005 & 0.0006 & 0.0005 & 0.0004 & 0.0004 & 0.0004 & 0.0005 & 0.0007 & 0.0006 \\
 & KNN & 0.0025 & 0.0131 & 0.0025 & 0.0025 & 0.0025 & 0.0025 & 0.0025 & 0.0030 & 0.0025 & 0.0025 & 0.0034 & 0.0027 & 0.0028 \\
 & LGBM & 0.0004 & 0.0006 & 0.0004 & 0.0007 & 0.0004 & 0.0004 & 0.0004 & 0.0007 & 0.0004 & 0.0004 & 0.0004 & 0.0005 & 0.0005 \\
 & LR & 0.0001 & 0.0001 & 0.0001 & 0.0001 & 0.0001 & 0.0001 & 0.0001 & 0.0001 & 0.0001 & 0.0001 & 0.0001 & 0.0001 & 0.0001 \\
 & MLP & 0.0001 & 0.0001 & 0.0001 & 0.0001 & 0.0001 & 0.0001 & 0.0001 & 0.0001 & 0.0001 & 0.0001 & 0.0001 & 0.0001 & 0.0001 \\
 & NB & 0.0001 & 0.0001 & 0.0001 & 0.0002 & 0.0001 & 0.0001 & 0.0001 & 0.0001 & 0.0001 & 0.0001 & 0.0001 & 0.0001 & 0.0001 \\
\end{longtable}
}
\begin{landscape}
{\footnotesize
\setlength{\tabcolsep}{2pt} 
\begin{longtable}{llrrrrrrrrrrrrr}
\caption{$R^2$ Score by dataset, model, and scaling method.} \label{tab:r2_big_table} \\
\toprule
Dataset & Model & NO & MM & MA & ZSN & PS & VAST & MC & RS & QT & DS & TT & LS & HT \\
\midrule
\endfirsthead
\caption[]{$R^2$ Score by dataset, model, and scaling method.} \\
\toprule
Dataset & Model & NO & MM & MA & ZSN & PS & VAST & MC & RS & QT & DS & TT & LS & HT \\
\midrule
\endhead
\midrule
\multicolumn{15}{r}{Continued on next page} \\
\midrule
\endfoot
\bottomrule
\endlastfoot
\multirow[t]{11}{*}{Abalone} & Ada & 0.1671 & 0.2210 & 0.2243 & 0.2503 & 0.1510 & 0.1938 & 0.1296 & 0.2301 & 0.1347 & 0.2018 & 0.1690 & 0.2112 & 0.2112 \\
 & CART & 0.1383 & 0.1464 & 0.1477 & 0.1560 & 0.1456 & 0.1500 & 0.1626 & 0.1417 & 0.1663 & 0.1416 & 0.1373 & 0.1686 & 0.1686 \\
 & CatBoost & 0.5226 & 0.5226 & 0.5225 & 0.5226 & 0.5225 & 0.5225 & 0.5225 & 0.5227 & 0.5224 & 0.5226 & 0.5226 & 0.5223 & 0.5223 \\
 & KNN & 0.5164 & 0.5023 & 0.4955 & 0.4662 & 0.5056 & 0.4406 & 0.5164 & 0.4552 & 0.4766 & 0.4100 & 0.4662 & 0.4699 & 0.4699 \\
 & LGBM & 0.5260 & 0.5278 & 0.5260 & 0.5256 & 0.5256 & 0.5256 & 0.5256 & 0.5190 & 0.5277 & 0.5260 & 0.5257 & 0.5257 & 0.5259 \\
 & LinearRegression & 0.5150 & 0.5150 & 0.5150 & 0.5150 & 0.5150 & 0.5150 & 0.5150 & 0.5150 & 0.4964 & 0.5150 & 0.5151 & 0.5204 & 0.5204 \\
 & MLP & 0.5245 & 0.5276 & 0.5265 & 0.5578 & 0.5580 & 0.5443 & 0.5559 & 0.5632 & 0.5434 & 0.5300 & 0.0226 & 0.5337 & 0.5560 \\
 & RF & 0.5244 & 0.5247 & 0.5249 & 0.5234 & 0.5244 & 0.5248 & 0.5247 & 0.5241 & 0.5230 & 0.5232 & 0.5239 & 0.5236 & 0.5236 \\
\end{longtable}
}
\end{landscape}
\begin{landscape}
{\footnotesize
\setlength{\tabcolsep}{2pt} 
\begin{longtable}{llrrrrrrrrrrrrr}
\caption{MSE by dataset, model, and scaling method.} \label{tab:mse_big_table} \\
\toprule
Dataset & Model & NO & MM & MA & ZSN & PS & VAST & MC & RS & QT & DS & TT & LS & HT \\
\midrule
\endfirsthead
\caption[]{MSE by dataset, model, and scaling method.} \\
\toprule
Dataset & Model & NO & MM & MA & ZSN & PS & VAST & MC & RS & QT & DS & TT & LS & HT \\
\midrule
\endhead
\midrule
\multicolumn{15}{r}{Continued on next page} \\
\midrule
\endfoot
\bottomrule
\endlastfoot
\multirow[t]{11}{*}{Abalone} & Ada & 8.4579 & 7.9106 & 7.8773 & 7.6131 & 8.6210 & 8.1869 & 8.8383 & 7.8177 & 8.7869 & 8.1054 & 8.4388 & 8.0102 & 8.0102 \\
 & CART & 8.7504 & 8.6683 & 8.6547 & 8.5702 & 8.6762 & 8.6316 & 8.5032 & 8.7161 & 8.4657 & 8.7169 & 8.7600 & 8.4426 & 8.4426 \\
 & CatBoost & 4.8477 & 4.8482 & 4.8485 & 4.8478 & 4.8486 & 4.8488 & 4.8492 & 4.8470 & 4.8501 & 4.8480 & 4.8477 & 4.8505 & 4.8506 \\
 & KNN & 4.9107 & 5.0541 & 5.1229 & 5.4200 & 5.0202 & 5.6801 & 4.9107 & 5.5324 & 5.3144 & 5.9916 & 5.4200 & 5.3829 & 5.3829 \\
 & LGBM & 4.8137 & 4.7948 & 4.8137 & 4.8170 & 4.8175 & 4.8175 & 4.8175 & 4.8846 & 4.7960 & 4.8137 & 4.8158 & 4.8158 & 4.8146 \\
 & LinearRegression & 4.9249 & 4.9249 & 4.9249 & 4.9249 & 4.9249 & 4.9249 & 4.9249 & 4.9249 & 5.1136 & 4.9249 & 4.9242 & 4.8700 & 4.8700 \\
 & MLP & 4.8285 & 4.7973 & 4.8087 & 4.4900 & 4.4885 & 4.6276 & 4.5099 & 4.4356 & 4.6370 & 4.7730 & 9.9246 & 4.7351 & 4.5088 \\
 & RF & 4.8295 & 4.8261 & 4.8244 & 4.8392 & 4.8299 & 4.8257 & 4.8268 & 4.8328 & 4.8436 & 4.8421 & 4.8342 & 4.8376 & 4.8376 \\
\end{longtable}
}
\end{landscape}
\begin{landscape}
{\footnotesize
\setlength{\tabcolsep}{2pt} 
\begin{longtable}{llrrrrrrrrrrrrr}
\caption{MAE by dataset, model, and scaling method.} \label{tab:mae_big_table} \\
\toprule
Dataset & Model & NO & MM & MA & ZSN & PS & VAST & MC & RS & QT & DS & TT & LS & HT \\
\midrule
\endfirsthead
\caption[]{MAE by dataset, model, and scaling method.} \\
\toprule
Dataset & Model & NO & MM & MA & ZSN & PS & VAST & MC & RS & QT & DS & TT & LS & HT \\
\midrule
\endhead
\midrule
\multicolumn{15}{r}{Continued on next page} \\
\midrule
\endfoot
\bottomrule
\endlastfoot
\multirow[t]{11}{*}{Abalone} & Ada & 2.5042 & 2.4037 & 2.3991 & 2.3437 & 2.5275 & 2.4467 & 2.5691 & 2.3881 & 2.5624 & 2.4383 & 2.4968 & 2.4173 & 2.4173 \\
 & CART & 2.0805 & 2.0702 & 2.0710 & 2.0582 & 2.0766 & 2.0702 & 2.0582 & 2.0718 & 2.0542 & 2.0789 & 2.0821 & 2.0502 & 2.0502 \\
 & CatBoost & 1.5612 & 1.5614 & 1.5612 & 1.5612 & 1.5617 & 1.5611 & 1.5616 & 1.5603 & 1.5620 & 1.5611 & 1.5609 & 1.5618 & 1.5618 \\
 & KNN & 1.5673 & 1.5938 & 1.6102 & 1.6555 & 1.5943 & 1.6962 & 1.5673 & 1.6654 & 1.6140 & 1.7400 & 1.6555 & 1.6416 & 1.6416 \\
 & LGBM & 1.5476 & 1.5486 & 1.5476 & 1.5500 & 1.5502 & 1.5502 & 1.5502 & 1.5632 & 1.5489 & 1.5476 & 1.5484 & 1.5484 & 1.5494 \\
 & LinearRegression & 1.6187 & 1.6187 & 1.6187 & 1.6187 & 1.6187 & 1.6187 & 1.6187 & 1.6187 & 1.6734 & 1.6187 & 1.6186 & 1.6132 & 1.6132 \\
 & MLP & 1.6159 & 1.5740 & 1.5748 & 1.5318 & 1.5059 & 1.5392 & 1.5258 & 1.4958 & 1.5429 & 1.5844 & 2.3339 & 1.5816 & 1.5126 \\
 & RF & 1.5590 & 1.5595 & 1.5584 & 1.5617 & 1.5596 & 1.5601 & 1.5593 & 1.5619 & 1.5613 & 1.5615 & 1.5622 & 1.5613 & 1.5613 \\
\end{longtable}
}
\end{landscape}
{\footnotesize
\setlength{\tabcolsep}{3pt} 
\begin{longtable}{llrrrrrrrrrrrrr}
\caption{Time to Train (s) by dataset, model, and scaling method.} \label{tab:train_reg_big_table} \\
\toprule
Dataset & Model & NO & MM & MA & ZSN & PS & VAST & MC & RS & QT & DS & TT & LS & HT \\
\midrule
\endfirsthead
\caption[]{Time to Train (s) by dataset, model, and scaling method.} \\
\toprule
Dataset & Model & NO & MM & MA & ZSN & PS & VAST & MC & RS & QT & DS & TT & LS & HT \\
\midrule
\endhead
\midrule
\multicolumn{15}{r}{Continued on next page} \\
\midrule
\endfoot
\bottomrule
\endlastfoot
\multirow[t]{11}{*}{Abalone} & Ada & 0.0971 & 0.1051 & 0.0774 & 0.0758 & 0.0961 & 0.0966 & 0.0967 & 0.0810 & 0.0971 & 0.0979 & 0.0973 & 0.1004 & 0.0978 \\
 & CART & 0.0095 & 0.0093 & 0.0093 & 0.0093 & 0.0094 & 0.0094 & 0.0093 & 0.0093 & 0.0093 & 0.0095 & 0.0094 & 0.0094 & 0.0094 \\
 & CatBoost & 0.6395 & 0.6556 & 0.6344 & 0.6431 & 0.6465 & 0.6445 & 0.6430 & 0.6503 & 0.6530 & 0.6478 & 0.6459 & 0.6400 & 0.6432 \\
 & KNN & 0.0010 & 0.0011 & 0.0010 & 0.0010 & 0.0010 & 0.0010 & 0.0010 & 0.0010 & 0.0010 & 0.0010 & 0.0011 & 0.0011 & 0.0011 \\
 & LGBM & 0.0286 & 0.0290 & 0.0288 & 0.0301 & 0.0303 & 0.0304 & 0.0304 & 0.0307 & 0.0293 & 0.0297 & 0.0292 & 0.0283 & 0.0303 \\
 & LinearRegression & 0.0003 & 0.0091 & 0.0004 & 0.0003 & 0.0004 & 0.0004 & 0.0003 & 0.0003 & 0.0003 & 0.0004 & 0.0004 & 0.0004 & 0.0004 \\
 & MLP & 0.8414 & 1.1040 & 1.1404 & 0.8932 & 1.1691 & 0.7039 & 1.5363 & 1.2343 & 1.8627 & 1.7507 & 0.3577 & 1.4888 & 1.5986 \\
 & RF & 0.5945 & 0.6241 & 0.6035 & 0.6003 & 0.5983 & 0.6006 & 0.5963 & 0.5977 & 0.6015 & 0.5973 & 0.6022 & 0.5991 & 0.5997 \\
\end{longtable}
}
{\footnotesize
\setlength{\tabcolsep}{3pt} 
\begin{longtable}{llrrrrrrrrrrrrr}
\caption{Time to Inference (s) by dataset, model, and scaling method.} \label{tab:inf_reg_big_table} \\
\toprule
Dataset & Model & NO & MM & MA & ZSN & PS & VAST & MC & RS & QT & DS & TT & LS & HT \\
\midrule
\endfirsthead
\caption[]{Time to Inference (s) by dataset, model, and scaling method.} \\
\toprule
Dataset & Model & NO & MM & MA & ZSN & PS & VAST & MC & RS & QT & DS & TT & LS & HT \\
\midrule
\endhead
\midrule
\multicolumn{15}{r}{Continued on next page} \\
\midrule
\endfoot
\bottomrule
\endlastfoot
\multirow[t]{11}{*}{Abalone} & Ada & 0.0025 & 0.0026 & 0.0019 & 0.0019 & 0.0025 & 0.0025 & 0.0025 & 0.0020 & 0.0025 & 0.0025 & 0.0025 & 0.0025 & 0.0025 \\
 & CART & 0.0002 & 0.0002 & 0.0002 & 0.0002 & 0.0002 & 0.0002 & 0.0001 & 0.0002 & 0.0001 & 0.0002 & 0.0002 & 0.0002 & 0.0002 \\
 & CatBoost & 0.0007 & 0.0008 & 0.0010 & 0.0009 & 0.0009 & 0.0010 & 0.0010 & 0.0008 & 0.0007 & 0.0009 & 0.0008 & 0.0008 & 0.0007 \\
 & KNN & 0.0034 & 0.0045 & 0.0048 & 0.0042 & 0.0045 & 0.0036 & 0.0035 & 0.0041 & 0.0042 & 0.0029 & 0.0044 & 0.0043 & 0.0044 \\
 & LGBM & 0.0006 & 0.0008 & 0.0007 & 0.0007 & 0.0007 & 0.0007 & 0.0007 & 0.0007 & 0.0007 & 0.0007 & 0.0007 & 0.0007 & 0.0007 \\
 & LinearRegression & 0.0000 & 0.0001 & 0.0000 & 0.0000 & 0.0000 & 0.0000 & 0.0000 & 0.0000 & 0.0000 & 0.0000 & 0.0000 & 0.0000 & 0.0000 \\
 & MLP & 0.0003 & 0.0003 & 0.0003 & 0.0003 & 0.0003 & 0.0003 & 0.0003 & 0.0003 & 0.0003 & 0.0003 & 0.0003 & 0.0003 & 0.0003 \\
 & RF & 0.0110 & 0.0107 & 0.0111 & 0.0110 & 0.0110 & 0.0110 & 0.0110 & 0.0109 & 0.0111 & 0.0109 & 0.0110 & 0.0111 & 0.0112 \\
\end{longtable}
}
{\footnotesize
\setlength{\tabcolsep}{1.2pt} 
\begin{longtable}{llrrrrrrrrrrrrr}
\caption{Memory Usage (kB) per Dataset and Scaling Method} \label{tab:memory_big_table} \\
\toprule
Dataset & NO & MM & MA & ZSN & PS & VAST & MC & RS & QT & DS & TT & LS & HT \\
\midrule
\endfirsthead
\caption[]{Memory Usage (kB) per Dataset and Scaling Method} \\
\toprule
Dataset & Model & NO & MM & MA & ZSN & PS & VAST & MC & RS & QT & DS & TT & LS & HT \\
\midrule
\endhead
\midrule
\multicolumn{14}{r}{Continued on next page} \\
\midrule
\endfoot
\bottomrule
\endlastfoot
\begin{tabular}[t]{@{}l@{}}Breast Cancer \\ Wisconsin \\ Diagnostic\end{tabular} & 0.1875 & 176.8210 & 175.7480 & 176.2767 & 229.9746 & 229.1289 & 228.2461 & 384.9801 & 267.6338 & 229.8695 & 232.2894 & 230.8722 & 232.3309 \\
Dry Bean & 1704.2923 & 2388.0706 & 2448.4049 & 2599.3600 & 2404.5947 & 2405.1895 & 2404.3014 & 2388.2383 & 2520.3968 & 2406.4144 & 2406.9097 & 2405.5592 & 2408.1296 \\
\begin{tabular}[t]{@{}l@{}}Glass \\ Identification\end{tabular} & 0.1875 & 23.2331 & 23.2064 & 26.1930 & 36.6759 & 36.0576 & 34.7992 & 51.7995 & 34.2393 & 36.1146 & 38.0356 & 37.9139 & 40.7885 \\
Heart Disease & 33.6901 & 49.0889 & 67.2025 & 71.7708 & 62.5277 & 62.8802 & 62.7256 & 121.9645 & 74.4512 & 63.1191 & 64.0603 & 64.0238 & 65.3331 \\
Iris & 0.1875 & 8.6296 & 8.7682 & 10.5488 & 16.9004 & 16.4834 & 15.6758 & 20.2601 & 22.0664 & 16.0218 & 18.4840 & 18.5913 & 20.3986 \\
\begin{tabular}[t]{@{}l@{}}Letter \\ Recognition\end{tabular} & 0.1875 & 2567.9333 & 3566.3210 & 3787.2116 & 3629.7077 & 3629.3434 & 3628.4954 & 2568.3698 & 3318.4468 & 3628.8434 & 3632.1798 & 3631.9978 & 3742.5224 \\
\begin{tabular}[t]{@{}l@{}}Magic Gamma \\ Telescope\end{tabular} & 0.1875 & 1552.4850 & 1552.0501 & 1552.2611 & 2198.9515 & 2199.2367 & 2198.2923 & 1552.7106 & 1792.9922 & 2198.5312 & 2201.3631 & 2201.1431 & 2306.3952 \\
\begin{tabular}[t]{@{}l@{}}Rice Cammeo \\ And Osmancik\end{tabular} & 211.2500 & 297.4365 & 358.2734 & 378.0465 & 304.2510 & 304.3102 & 303.1006 & 297.6322 & 359.7692 & 304.2708 & 305.9161 & 305.7961 & 328.2706 \\
\end{longtable}
}